\documentclass{article}

\usepackage{arxiv}
\usepackage[utf8]{inputenc} % allow utf-8 input
\usepackage[T1]{fontenc}    % use 8-bit T1 fonts
\usepackage{hyperref}       % hyperlinks
\usepackage{url}            % simple URL typesetting
\usepackage{booktabs}       % professional-quality tables
\usepackage{amsfonts}       % blackboard math symbols
\usepackage{nicefrac}       % compact symbols for 1/2, etc.
\usepackage{microtype}      % microtypography
\usepackage{lipsum}		% Can be removed after putting your text content

\usepackage{graphicx}     
\usepackage{subcaption}
\usepackage{amsmath}

\title{A connection between the pattern classification problem and the General Linear Model for statistical inference}

\author{
  J.M. Gorriz\thanks{gorriz@ugr.es, jg528@cam.ac.uk} \\
  Department of Signal Theory and Communications\\
  University of Granada\\
  Granada, Spain \\
   \And
 SiPBA Group \\
  Department of Signal Theory and Communications\\
  University of Granada\\
  Granada, Spain\\
  \texttt{sipba@ugr.es} \\
     \And
 John Suckling \\
  Department of Psychiatry\\
  University of Cambridge\\
  Cambridge, UK\\
  \texttt{js369@cam.ac.uk} \\
       \And
       International Initiatives\\
for the Alzheimer's Disease Neuroimaging Initiative (ADNI)\\
for the Parkinson’s Progression Markers Initiative (PPMI)\\
}

\begin{document}
\maketitle

\begin{abstract}
A connection between the General Linear Model (GLM) in combination with classical statistical inference and the machine learning (MLE)-based inference is described in this paper. Firstly, the estimation of the GLM parameters is expressed as a Linear Regression Model (LRM) of an indicator matrix, that is, in terms of the inverse problem of regressing the observations. In other words, both approaches, i.e. GLM and LRM, apply to different domains, the observation and the label domains, and are linked by a normalization value at the least-squares solution. Subsequently, from this relationship we derive a statistical test based on a more refined predictive algorithm, i.e. the (non)linear Support Vector Machine (SVM) that maximizes the class margin of separation, within a permutation analysis. The MLE-based inference employs a residual score and includes the upper bound to compute a better estimation of the actual (real) error. Experimental results demonstrate how the parameter estimations derived from each model resulted in different classification performances in the equivalent inverse problem.  Moreover, using real data the aforementioned predictive algorithms within permutation tests, including such model-free estimators, are able to provide a good trade-off between type I error and statistical power.
\end{abstract}

% keywords can be removed

\section{Introduction}

Nowadays, there is an open question about the usefulness of machine learning (MLE) techniques to test significance of group analyses.  While classification problems using MLE have been the main target in predictive or decoding analysis in neuroimaging, there is an increasing interest in the inference analysis with continuous outputs based on MLE, as detailed in \cite{Cohen2011} and remarked in \cite{Reiss15}.

Recently, several advances for combining p-value maps have been proposed based on the concept of \emph{prevalence} \cite{Heller07,Rosenblatt14}, beyond the fixed and mixed (random) effects models \cite{Friston02}. Common to all these approaches is to assume a voxel-wise model that allows a proportion of conditions or subjects that activated the voxel at some mixing proportion.  This assumption, that is more realistic than those assumed in classic random effect approaches, e.g. homogeneity in the (binary) activation pattern \cite{Rosenblatt14}, clearly opens a new application field for modern statistics.  

Indeed, the concept of prevalence as a fraction of individuals correctly classified by MLE algorithms in group comparisons, is not novel at all in neuroimaging, being the main focus of predictive inference. As an example, out-of sample generalization approaches, such as Cross-Validation (CV), try to estimate on unseen new data the accuracy ($A_{cc}$) of the classifier in a binary classification problem. Despite the methods and goals of predictive CV inference being distinct from classical extrapolation procedures \cite{Lindquist13}, they are actually exploited within statistical frameworks aimed at assessing statistical significance \cite{Reiss15}. Examples include bootstrapping, binomial or permutation (``resampling'') tests \cite{Winkler16}, which have been demonstrated to be competitive outside the comfort zone of classical statistics, filling otherwise-unmet inferential needs. 

In the pattern classification problem we usually assume the existence of classes ($H_1$) that are differentiated by classifiers that are measured by their performance in terms of $A_{cc}$ or \emph{prevalence} on a independent dataset. Then, we conclude (improperly in a statistical sense) $H_1$ using empirical confidence intervals, e.g. standard deviations of the classification $A_{cc}$ from training folds. In limited sample sizes the most popular K-fold CV method \cite{Kohavi95} has been demonstrated to sub-optimally work under unstable conditions \cite{Gorriz18,Gorriz19,Varoquaux18}. In such circumstances, the predictive power of the fitted classifiers can be arguable. Moreover, recent works have partially demonstrate that, when using only a classifier's empirical $A_{cc}$ as a test statistic, the probability of detecting differences between two distributions is lower than that of a bona fide statistical test \cite{Rosenblatt16,Kim20}.

Beyond the latter empirical techniques for the estimation of performance, MLE is well-framed into a data-driven statistical learning theory (SLT) which is mainly devoted to problems of estimating dependencies with limited amounts of data \cite{Vapnik82}. Although CV-MLE approaches were not originally designed to test hypotheses based on prevalence in brain mapping \cite{Friston2013}, they are theoretically grounded to provide confidence intervals in the classification of image patterns (protected inference) that can be seen as maps of statistical significance \cite{Gorriz2021}. As shown in the latter reference,  this can be achieved by assessing the upper bounds of the actual error in a binary classification problem (a confidence interval), and by using simple significance tests of a population proportion within it.  Definitely, this results in improvements to the test's statistical power based on $A_{cc}$. Thus, assessing with high probability the quality of the fitting function (and its generalization ability) in terms of in and out-sample predictions can be conceptualized, under a hypothesis testing scenario, as the inverse problem of ``carefully rejecting $H_0$'', that is, the problem of rejecting $H_1$, and thus accepting $H_0$ (there is no effect or it is not significant).

In this paper we show the connection between the classical General linear model (GLM) including the classic random effect model, with the MLE framework in the estimation of model/classifier parameters, and the subsequent analyses to achieve the degree of significance in group comparisons.  In this sense, inference based on the parametric T statistic or the prevalence-based probability tests, means two different paths to assess the same problem.  Moreover, we show a novel method for achieving statistical significance using MLE and permutation tests based on concentration inequalities. This approach assesses the worst case of the actual error to propose an estimation of the observed distribution of the permuted data.

\section{Methods: Classical and MLE statistical inferences}\label{sec:methods}

\subsection{Background on classical statistics in neuroimaging}\label{sec:GLM}

The GLM \cite{Friston02} is defined just for a single observation level, e.g.  in a inter-subject comparison, as:
\begin{equation}\label{eq:1}
\mathbf{y}=\mathbf{X} \mathbf{\theta} + \mathbf{\epsilon}
\end{equation}
where $\mathbf{y}$ is the $N\times 1$ observation vector with units over time, voxels, etc., $\mathbf{\epsilon}$ is the $N\times 1$ vector of errors that is assumed to be Gaussian, $\mathbf{X}$ is the $N \times M$ matrix containing the explanatory variables or constraints and $\mathbf{\theta}$ is the $M\times 1$ vector of parameters explaining the observations $\mathbf{y}$. Note that i) for a hierarchical observation model each level like the latter requires the estimation of the previous levels, and ii) in terms of MLE, $\mathbf{X}$ plays the role of a multidimensional label or regressors acting on the observations $\mathbf{y}$.  In the classic GLM $\mathbf{\theta}$ is usually estimated by a Maximum Likelihood (ML) criterion based on the Gaussian assumption and is given by:
\begin{equation}\label{eq:2}
\hat{\mathbf{\theta}}=(\mathbf{X}^T \mathbf{C}_{\epsilon}^{-1} \mathbf{X})^{-1}\mathbf{X}^T \mathbf{C}_{\epsilon}^{-1}\mathbf{y}
\end{equation}
Inferences about this estimate that is, how large are the components of $\mathbf{\theta}$ and their relation with each other, can be obtained by using a linear compound, specified by a contrast weight vector $\mathbf{c}$, and writing a T statistic as:
\begin{equation}\label{eq:3}
T=\frac{\mathbf{c}^T\hat{\mathbf{\theta}} }{\sqrt{\mathbf{c}^TCov(\hat{\mathbf{\theta}})\mathbf{c} }}
\end{equation}
where $Cov(\hat{\mathbf{\theta}})=(\mathbf{X}^T \mathbf{C}_{\epsilon}^{-1} \mathbf{X})^{-1}$. The T statistic gives us the probability of observation of the ML estimation under the null hypothesis and, when it is small enough, e.g.  $p<0.05$, the linear compound is considered significantly different from zero. As an example, given a set of two parameters in $\theta=[\theta_1, \theta_2]^T$ if we select $\mathbf{c}=[1 -1]$ we are assessing how large is the first parameter w.r.t the second, i.e. the difference  $\theta_1-\theta_2$, thus if the T statistic provides a small probability, the latter difference is statistically significant and observations are generated from different sources. 

A similar procedure could be established based on a Bayesian estimation and inference to handle complex hierarchical observation models. The latter framework is based on the Expectation Maximization (EM) algorithm for parameter estimation and known priors and a-priori probability models, with the aim of evaluating the posterior probability (ppm). By thresholding the ppm, a relationship between the two approaches can be established including similarities (statistical power) and differences (specificity) \cite{Friston02}.

\subsubsection{Least Squares of the GLM}
\label{sec:GLMleast}
The GLM can be estimated without any assumption on the noise model by simply solving the associated Least Squares (LS) problem. Therefore, if we assume that $\epsilon=0$ in the GLM, the problem is now to find the ``best'' set of parameters $\theta_i$ that better explains each observation $y_i$ by:
\begin{equation}\label{eq:4}
y_j=\sum_{i=1}^M X_{ji} \theta_i;\quad \text{for } j=1,\ldots,N
\end{equation}
Thus, we need to solve the linear regression problem given in equation \ref{eq:4} to estimate the parameters $\theta_i$. The most popular estimation method is LS, in which we select the coefficients $\mathbf{\theta}$ to minimize the residual sum of squares:
\begin{equation}\label{eq:5}
RS(\mathbf{\theta})=\sum_{j=1}^N (y_j-\sum_{i=1}^M X_{ji} \theta_i)^2
\end{equation}
The solution to this problem ($\frac{\partial RS(\mathbf{\theta})}{\partial \mathbf{\theta}}=0$), the Markov-Gauss estimate, provides the smallest variance among all linear unbiased estimates and is given by:
\begin{equation}\label{eq:6}
\hat{\mathbf{\theta}}=(\mathbf{X}^T\mathbf{X})^{-1}\mathbf{X}^{T}\mathbf{y}
\end{equation}
similar to equation \ref{eq:2} in the GLM but assuming $C_{\epsilon}=\mathbf{I}$, that is, the errors are assumed to be independently and identically distributed.

\subsection{Converting the problem of the estimation of $\theta$ into a LS classification problem}\label{sec:LS}

In the LS multiclass classification problem, the goal is to design $M$ linear functions $f_i(\mathbf{y})=\mathbf{w}_i^T\mathbf{y}$, given a set of response variables $\mathbf{y}_i$ and according to a suitable mean squared error (MSE) criterion w.r.t some desired discrete output (i.e. labels) $\mathbf{x}_i$ which represents a binary code. Note how, in general, this pattern classification problem is the inverse problem of hierarchical modeling in neuroimaging, as shown in the following. Recently, the residual score or the classification error obtained from several methodologies beyond LS, e.g. by applying the fitted linear hyperplanes to new unseen data, are deployed to establish a CV $A_{cc}$-based test on the responses with permuted labels, as shown elsewhere \cite{Reiss15, Gorriz2021}.

\subsubsection{The inverse problem: LS for regressing an indicator matrix}\label{sec:LSinverse}
Let consider the general inverse problem, that is, given a set of observations $\{\mathbf{y}_i\}$, for $i=1,\ldots,N$, we are interested in explaining a set of ``explanatory'' binary-coded variables $\mathbf{x}_i$ (labels) by a matrix $\mathbf{W}$ of parameters. This problem that is referred in this paper as the inverse problem in the \emph{label domain}, is also known as the linear regression of an Indicator Matrix or linear regression model (LRM) \cite{Hastie2001}. In this model we regress explanatory variables instead of doing that on the observed responses, i.e. in the \emph{observation domain}, as in the GLM. This regression could be more accurate than the latter depending on the nature of the data to be fitted, e.g. for a low number of discrete categories in the specified design matrix $\mathbf{X}=[x_{im}]$.

If we have $M$ classes then $\mathbf{X}$ is a $N\times M$ matrix, where each row $i=1,\ldots,N$ contains a single $x_{im}=1$, for $m=1,\ldots,M$,  $\mathbf{Y}$ is the $N\times P$ matrix of column responses $\mathbf{y}_i$ and $\mathbf{W}$ is a $P\times M$ coefficient matrix. Thus, we fit a linear regression model of the form:
\begin{equation}\label{eq:7}
\mathbf{X}=\mathbf{Y}\mathbf{W}
\end{equation}
where the $P$ dimension allows the inclusion of several responses (multimodality or multiframe acquisitions) given the same indicator response matrix $\mathbf{X}$. Following the same methodology as aforementioned the best estimation is given by:
\begin{equation}\label{eq:8}
\hat{\mathbf{W}}=(\mathbf{Y}^T\mathbf{Y})^{-1}\mathbf{Y}^T\mathbf{X}
\end{equation}
which regresses inputs of observations into a novel set of labels or constraints:
\begin{equation}\label{eq:9}
\hat{\mathbf{X}}=\mathbf{Y}\hat{\mathbf{W}}
\end{equation}
The novel set $\hat{\mathbf{X}}$ can be seen as a guess on the constraints for the set of observation vectors $\mathbf{y}_i$, or an approximation of the posterior probability $p(\text{class}=m|\mathbf{y})$. Thus, it allows us to compute an error model as:
\begin{equation}\label{eq:10}
\epsilon_{LS}=\mathbf{X}-\hat{\mathbf{X}}
\end{equation}

\subsubsection{Connection between $\theta$ and $\mathbf{w}$}
\label{sec:connection}
For simplicity and to connect with the GLM as shown in section \ref{sec:GLM}, let $P=1$ in the LRM, then $\mathbf{W}=\mathbf{w}$ is a $1\times M$ row vector and $\mathbf{Y}=\mathbf{y}$ is an $N\times 1$ column vector. An easy relation between the GLM and LRM approximations can be found taking into account that:
\begin{equation}\label{eq:11}
\mathbf{X}=\mathbf{y}\mathbf{w}+\mathbf{\epsilon}_{LS}
\end{equation}
Thus, the corresponding GLM is:
\begin{equation}\label{eq:12}
\mathbf{y}=(\mathbf{X}-\mathbf{\epsilon}_{LS})\hat{\theta}
\end{equation}
where we define $\hat{\theta}=\mathbf{w}^T(\mathbf{w}\mathbf{w}^T)^{-1}$ and the GLM noise model is derived using $\epsilon=-\epsilon_{LS}\hat{\theta}$. The scalar term of equation \ref{eq:12} can be expressed at the LS solution as:
\begin{equation}\label{eq:13}
(\mathbf{w}\mathbf{w}^T)^{-1}=(\mathbf{y}^T\mathbf{y})^2/((\mathbf{X}^T\mathbf{y})^T\mathbf{X}^T\mathbf{y})=\frac{(\sum_{i=1}^N y_i^2)^2}{\sum_{m=1}^M\sum_{i,j}y_{im}y_{jm}}
\end{equation}
where $y_{im}$ denotes the observation $i$ belonging to class $m$. Thus, the LS linear regression of the observations can be described by the GLM on the observations (a linear regression on the explanatory variables) and viceversa.\footnote{given a GLM on the observation we can define a LRM on the explanatory variables as $\hat{\mathbf{w}}=\mathbf{\theta}^T(\mathbf{\theta}\mathbf{\theta}^T)^{-1}$ with and error $\epsilon_{LS}=-\epsilon\hat{\mathbf{w}}$}

\subsubsection{Inference of the inverse GLM based on MLE}
\label{sec:inference}
The LRM can be seen as a generalization of the GLM for the responses, coding $\mathbf{x}$ as a vector of continuous noisy responses (then $M=1$,  instead of being an indicator matrix):
\begin{equation}\label{eq:14}
\mathbf{x}=\mathbf{Y}\mathbf{w}+\mathbf{\epsilon}
\end{equation}
that is equivalent to the inverse GLM in equation \ref{eq:1}. An inference on this model based on MLE could proceed as follows.  Given an a set of pairs $(\mathbf{y}_i,x_i)$ we estimate the set of parameters $\mathbf{w}$ using a similar expression as in equation \ref{eq:8}. After the fitting process, we assess its significance under the null hypothesis on an independent set, likewise the T-statistic inference on the GLM, using a CV $A_{cc}$-based test statistic as:  
\begin{equation}\label{eq:15}
T_{CV}=(\mathbf{x}-\mathbf{Y}\mathbf{w})^T(\mathbf{x}-\mathbf{Y}\mathbf{w})=\sum_{i=1}^N(x_i-\sum_{j=1}^PY_{ij}w_j)^2
\end{equation}
The null distribution is modeled by choosing a large number of permutations $\pi$ to create artificial data sets, $(\mathbf{y}_i,x_{\pi_p})$, for $p=1,\ldots, O$, i.e. a permutation test, and evaluating the sum of squared residuals $T_{CV}$ on every unseen sample within the permuted and original set. Consequently, the p-value is defined by\footnote{the correction factor +1  in the numerator and denominator is justified by the inclusion of the original sample set in the test}:
\begin{equation}\label{eq:16}
p_{value}=\frac{ card\{ T_{CV}^{\pi}<T_{CV}\}+1}{O+1}
\end{equation}
where $card(.)$ is the cardinality of a set and $T_{CV}$ and $T_{CV}^{\pi}$ are the CV $A_{cc}$-based tests on the original and permuted sets, respectively.

In the latter test, also known as P-test \cite{Reiss15}, we assumed that we have a good procedure for estimating $\mathbf{w}$. However, CV is a standard procedure for estimating the actual error of the classifier that is found to be unstable in limited samples sizes \cite{Varoquaux18, Gorriz19}. Thus, we could improve this estimation by including a term to cope with the possibility that the fitting process is not as good as expected, i.e. the resulting estimate is not a good predictor. In this sense other alternatives could be tested by the assessment of the worst case based on concentration inequalities and the resubstitution estimate as:
\begin{equation}\label{eq:17}
T_{Res}=(\mathbf{x}-\mathbf{Y}\mathbf{w})^T(\mathbf{x}-\mathbf{Y}\mathbf{w})+\Delta(N,P)
\end{equation}
where $\Delta(N,P)$ is an upper bound of the actual risk \cite{Vapnik82, Gorriz2021,Gorriz19} with a probability at least $1-\alpha$.

\subsection{A general framework for multiclass regression}

As pointed out the columns in the specified design matrix $\mathbf{X}$ can be interpreted and coded as the labels of a multiclass classification problem for each component of the observation variable or response $\mathbf{y}$, stored in the rows of $\mathbf{Y}$. Let $\mathbf{x}$ denote the vector of labels for each observation $\mathbf{y}$, then it is straight forward to see that the following minimization is similar to the one proposed above (eq. \ref{eq:5}) for estimation of parameters contained in $\theta$:
\begin{equation}\label{eq:18}
\begin{array}{l}
\mathbf{W}=arg\min_{\mathbf{W}}E[||\mathbf{X}-\mathbf{Y}\mathbf{W}||^2]=arg\min_{\mathbf{W}}E[\sum_{m=1}^M (\mathbf{x}_{m}-\mathbf{Y}\mathbf{w}_m)^2]=\\=
arg\min_{\mathbf{W}} \sum_{m=1}^M \sum_{j=1}^N (x_{jm}-\sum_{i=1}^P y_{ji}w_{im})^2
\end{array}
\end{equation}
where $m,j,i$ index the set of labels, samples and observations, respectively, $\mathbf{W}=(\mathbf{w}_1,\ldots,\mathbf{w}_M)^T$ is the matrix of linear functions and $\mathbf{x}_j$ are the set of constraints (labels) for each observation $\mathbf{y}_j$. This is equivalent to solve $M$ LS problems one for each class. It is worth mentioning that in the GLM the parameters $\mathbf{\theta}$ and observations $\mathbf{y}$ are considered as vectors ($P=1$). 

In previous sections we have shown a simple connection between GLM and LRM, although the goal of this work is not using the parameters derived from LRM,  at all. LRM is a naive LS model based on the minimization of the empirical risk, e.g. the MSE. In this sense, we prefer to use the Structural Risk Minimization (SRM) principle by means of support vector machines (SVM), an optimum strategy with limited sample sizes. The latter minimization is more related to the concentration inequalities framework as shown in \cite{Vapnik82}. Thus, linear regression could be replaced by SVM or other predictive algorithms, that employ different loss functions and measures of performance as suggested in \cite{Reiss15}. In this sense it would be interesting to assess the connection between the parameters estimated by SVM with the ones obtained using GLM. In fact, for the linear SVM, after training, the same equations, i.e. equation \ref{eq:12}, could be applied to estimate $\theta_{SVM}$ in terms of the support vectors, as we did with the LRM (see experimental section below).

\section{Experimental results}

In the first part of the experiments, to clearly state the problem and the solutions, we consider a simple group comparison with only one-level analysis (a Bayesian approach of this problem is equivalent to the GLM based inference on this single level) using a second-level design matrix that models the subject-specific effects over subjects. This is the well-known example in diagnostics in the group comparison of two conditions, e.g. Alzheimer subjects vs controls. We adjust the GLM and the equivalent problem using LRM and SVM. Thus, we regress observed variables using a simple explanatory matrix $\mathbf{X}$ and a Gaussian model for the noise to obtain two parameters $\theta_1,\theta_2$ in this toy example, as the following:
\begin{equation*}\label{eq:19}
\mathbf{Y}|_{N\times 1}=\mathbf{X}|_{N\times 2} \mathbf{\theta}|_{2\times 1}+\mathbf{\epsilon}|_{N\times 1}
\end{equation*}
where, as an example, 
\begin{equation*}\label{eq:20}
\mathbf{X}=\left(
\begin{array}{cc}
1 &0 \\
0 &1 \\
1 &0 \\
1 &0\\ 
...&...\\
\end{array}\right)
\end{equation*}
is a matrix of explanatory variables containing $1$s and $0$s and indicates the class of the observation using a two dimensional binary code. A hierarchical model could be processed the same way by fitting the set of parameters step by step, however we are interested in assessing the connection between $\mathbf{\theta}$ and $\mathbf{w}$ in this toy example. The objective of this part is double: i) the estimation of model parameters using both methodologies and domains linking them by the use of the theoretical connection in equation \ref{eq:12} and, ii) to assess how well they explain observations and labels in both domains. The latter can be tackled by showing the estimations and the group of observations in both domains and by quantitatively evaluating the classification error in the equivalent label domain, given the expected ideal values for model parameters.

In the last part of this section, we show the inference analysis derived from the two methodologies on each domain.  We regress on the observations and on the labels to construct and assess the spatially extended statistical processes, which provide maps of significance, using the MRI ADNI dataset \cite{Gorriz2021}. In this way, we compare SPM, that is based on a two-sample T-statistic similar to equation \ref{eq:3}, where significance is individually assessed at each voxel with a using three configurations: cluster-defining threshold CDT of $P = 0.001$ (uncorrected for multiple comparisons), cluster extent threshold equal to $10$ and FWE correction at $0.05$, and the P-tests described in section \ref{sec:inference}.

\subsection{Data generation 1}

A $N$-dimensional Gaussian noise vector $\mathbf{v}$ is randomly drawn with zero mean and an $N\times N$ covariance matrix with $2$-norm equal to $1$.  This noise allows the definition of a vector of observations by adding the noise to a binary vector (a column in the explanatory matrix of indicators), i.e.  $\mathbf{y}=\mathbf{X}_k+\mathbf{v}$ for $k\in\{1,2\}$.  The design matrix is then obtained by $X=[\mathbf{X}_k \bar{\mathbf{X}}_k]$,  where $\bar{(.)} $ denotes logical negation.

\begin{figure*}
\centering
\includegraphics[width=\textwidth]{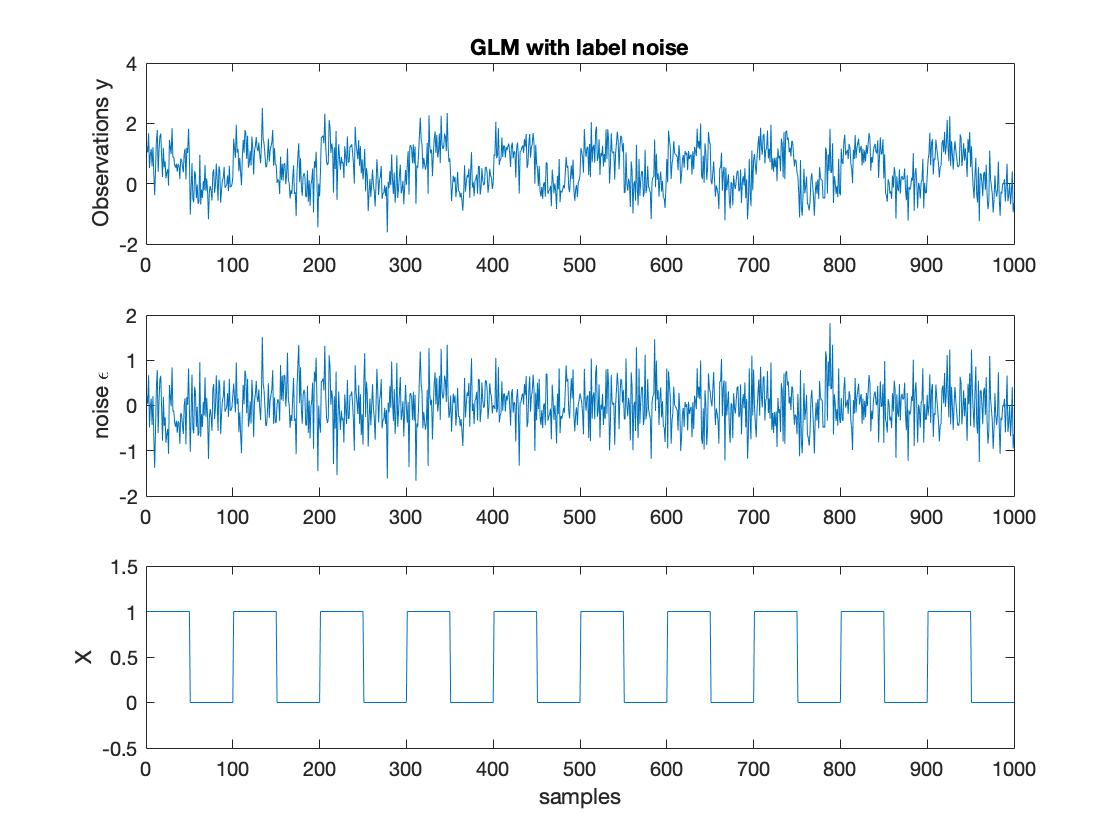}
\caption{Data generation 1 (DG1) example}
\label{fig:uno}
\end{figure*}

\begin{figure*}
\centering
\includegraphics[width=\textwidth]{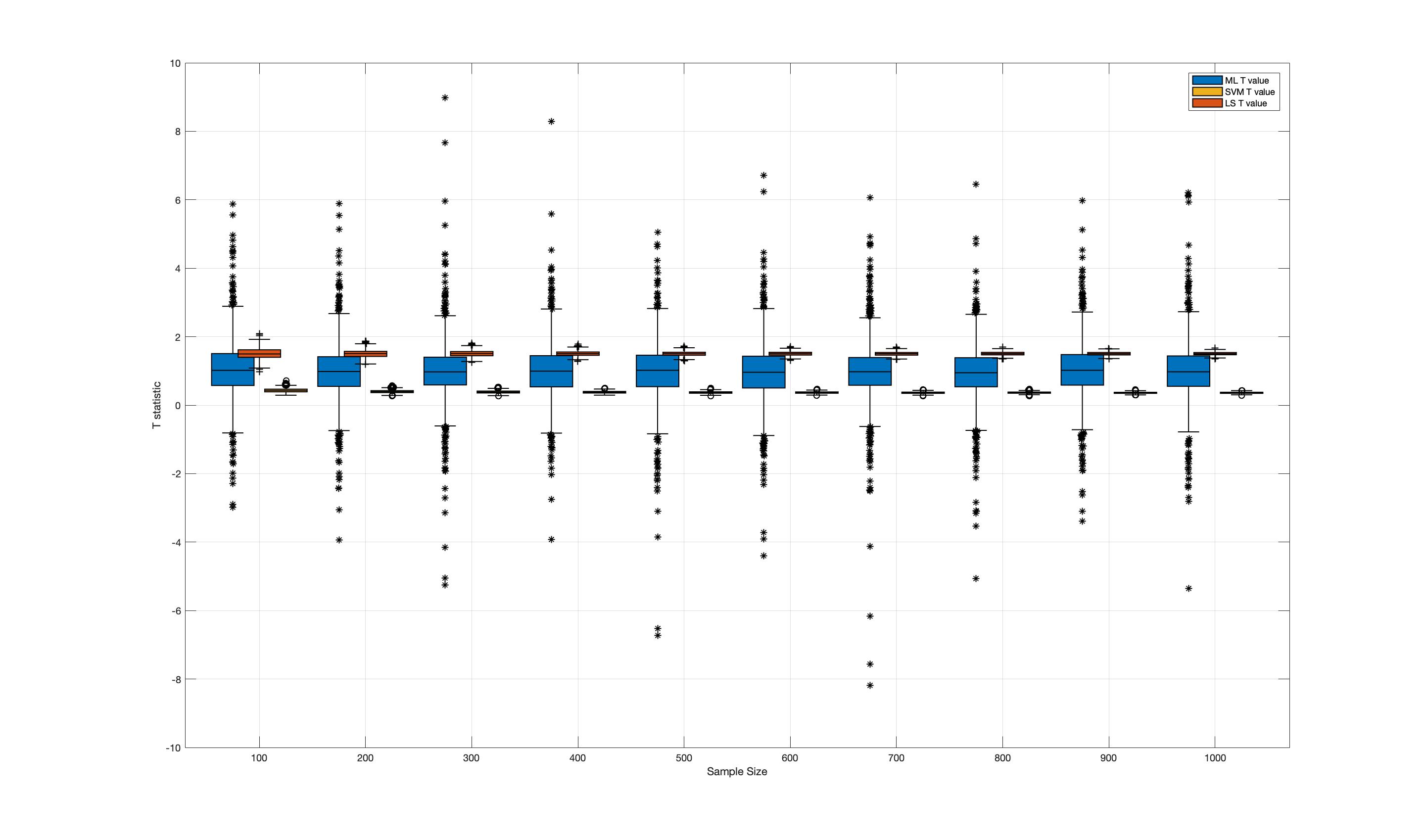}
\includegraphics[width=0.49\textwidth]{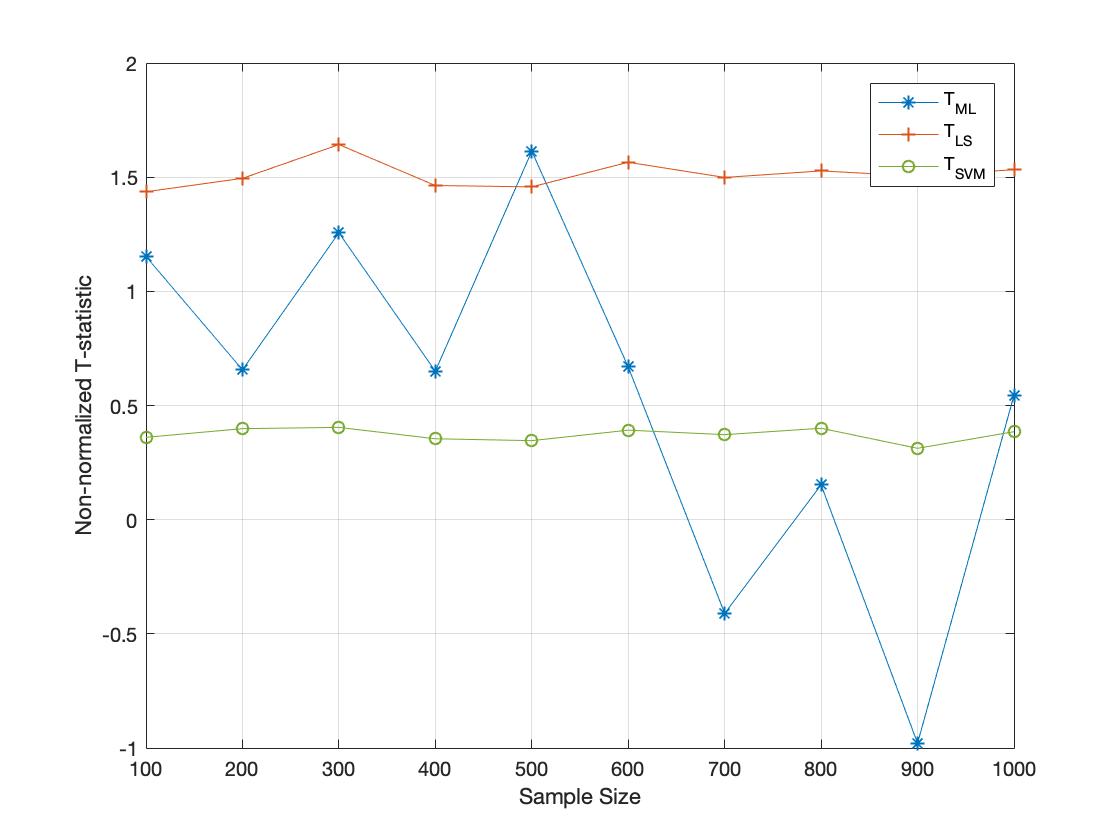}
\includegraphics[width=0.49\textwidth]{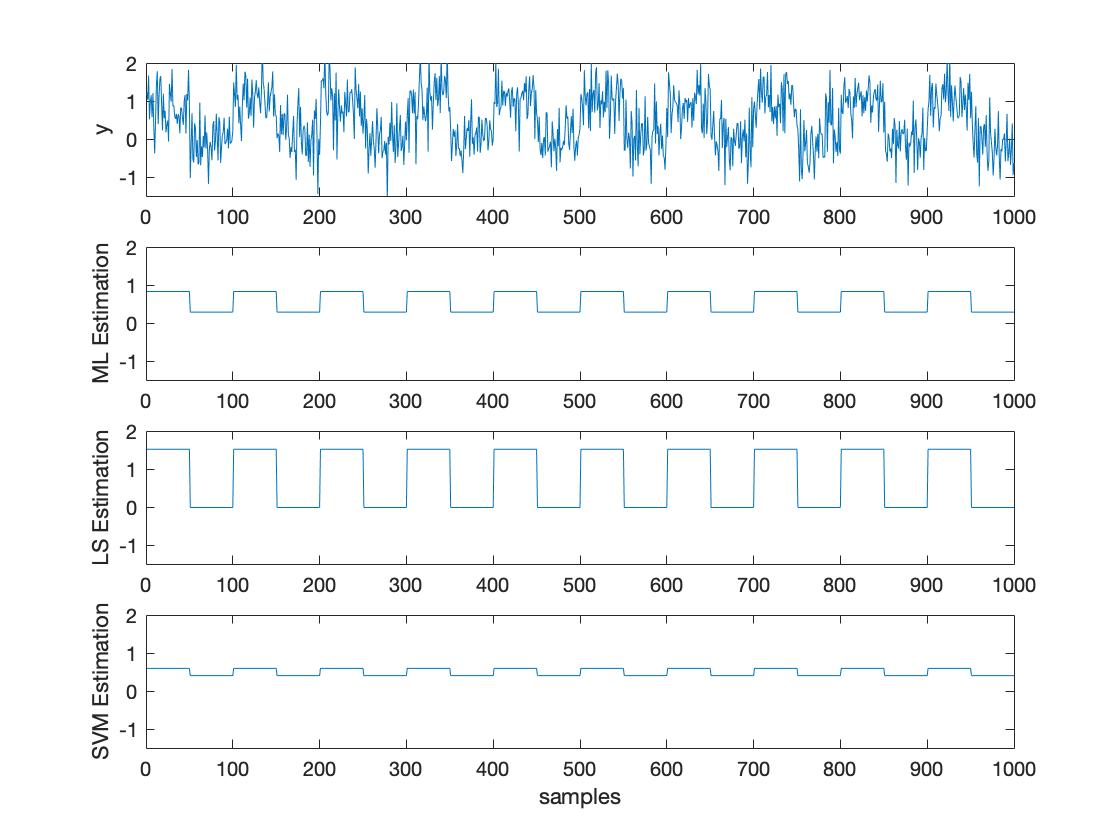}
\caption{Estimated Observations and T-statistic distribution.  Note that in the GLM model we use the covariance matrix of the noise to evaluate \ref{eq:2}, that is, in the estimation of $\theta$.  We show the comparison between non-normalized statistics of all the estimations, that is,  suppressing the covariance term in the GLM,  in a random (R=1000) simulation. This clearly demonstrates that only on average the ML statistic converges to the ideal value $\theta_1-\theta_0=1$ unlike the single sample of this distribution shown in the bottom left.}
\label{fig:dos}
\end{figure*}

Once the observations are artificially drawn (see figure \ref{fig:uno}) with increasing sample size, we regress both explanatory variables (LS or SVM) and observations (GLM) to obtain a set of two parameters for each model $\mathbf{\theta}=[\theta_1,\theta_2]$, $\mathbf{w}=[w_1,w_2]$.  All the methods can be employed to estimate the regressed observation variables using equations \ref{eq:1} and \ref{eq:10}, given the explanatory matrix and the estimated parameters, as shown in figure \ref{fig:dos}. In the latter we plot the distribution of the T-statistic over 1000 simulations (up), a sample of this distribution that shows the variability of the estimation using GLM around the ideal value 1 (bottom left) and the estimated observation by the analyzed models.  

\begin{figure*}
\centering
\includegraphics[width=\textwidth]{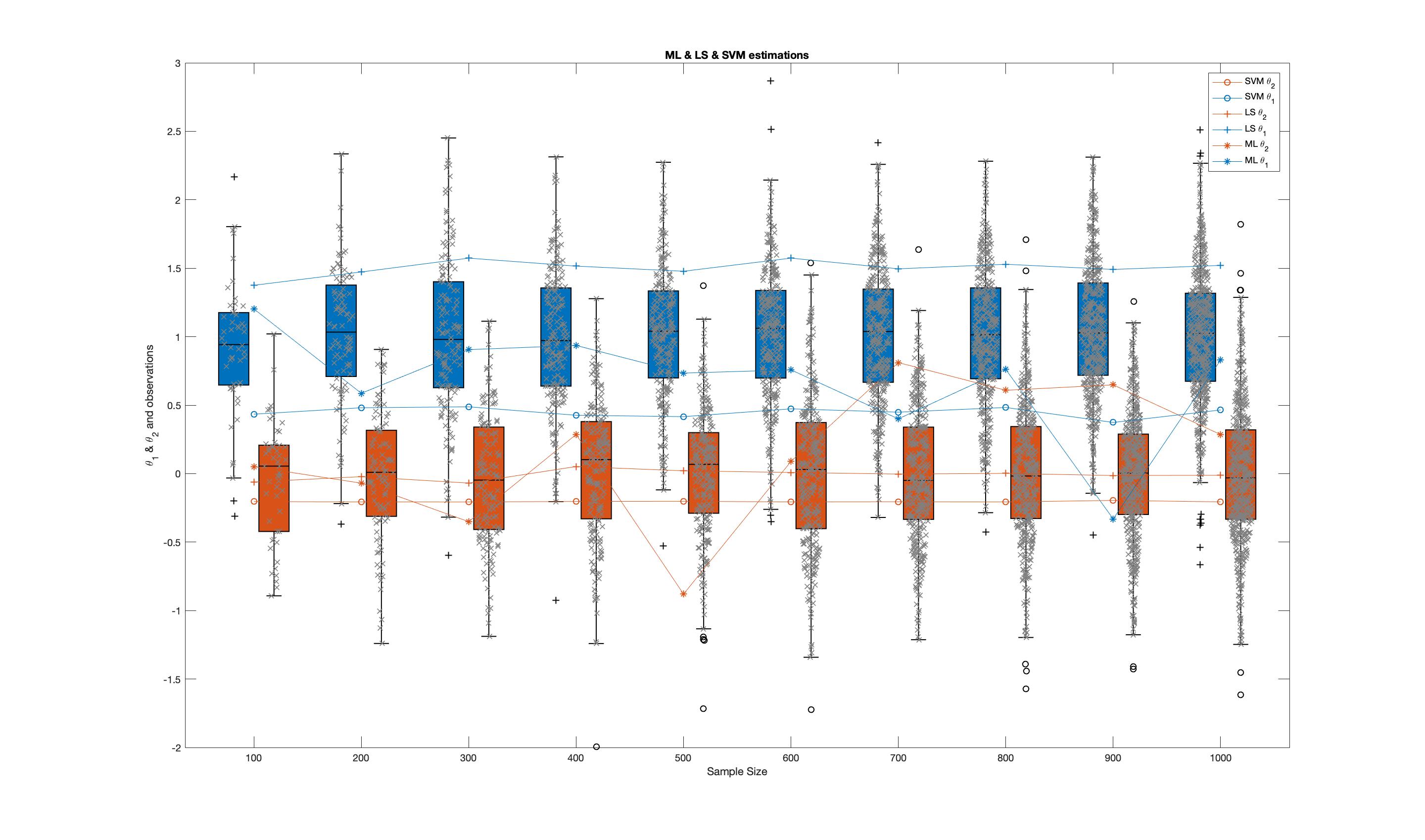}
\caption{Distribution of observations and estimations of  $\mathbf{\theta}$ for GLM,  LRM and SVM in DG1}
\label{fig:tres}
\end{figure*}

\begin{figure*}
\centering
\includegraphics[width=\textwidth]{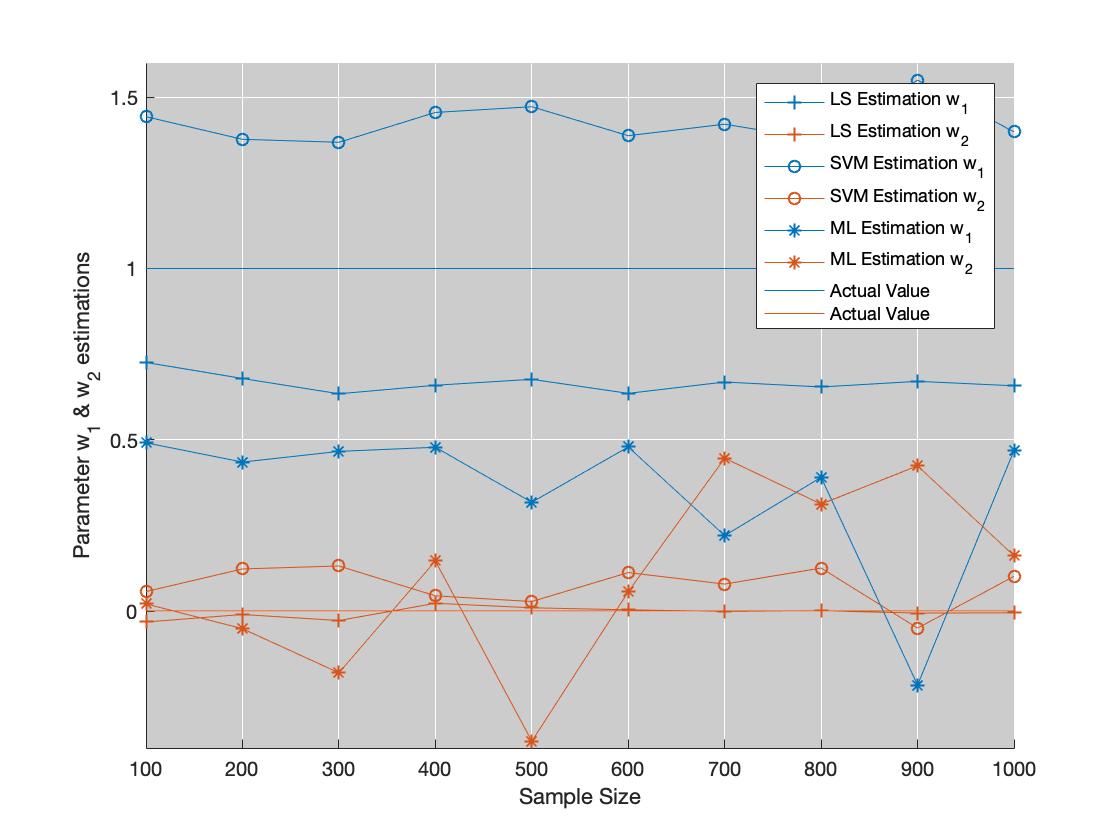}
\caption{Estimations of the parameter $\mathbf{w}$ regressing the observations with increasing sample size in DG1}
\label{fig:cuatro}
\end{figure*}

\begin{figure*}
\centering
\includegraphics[width=0.49\textwidth]{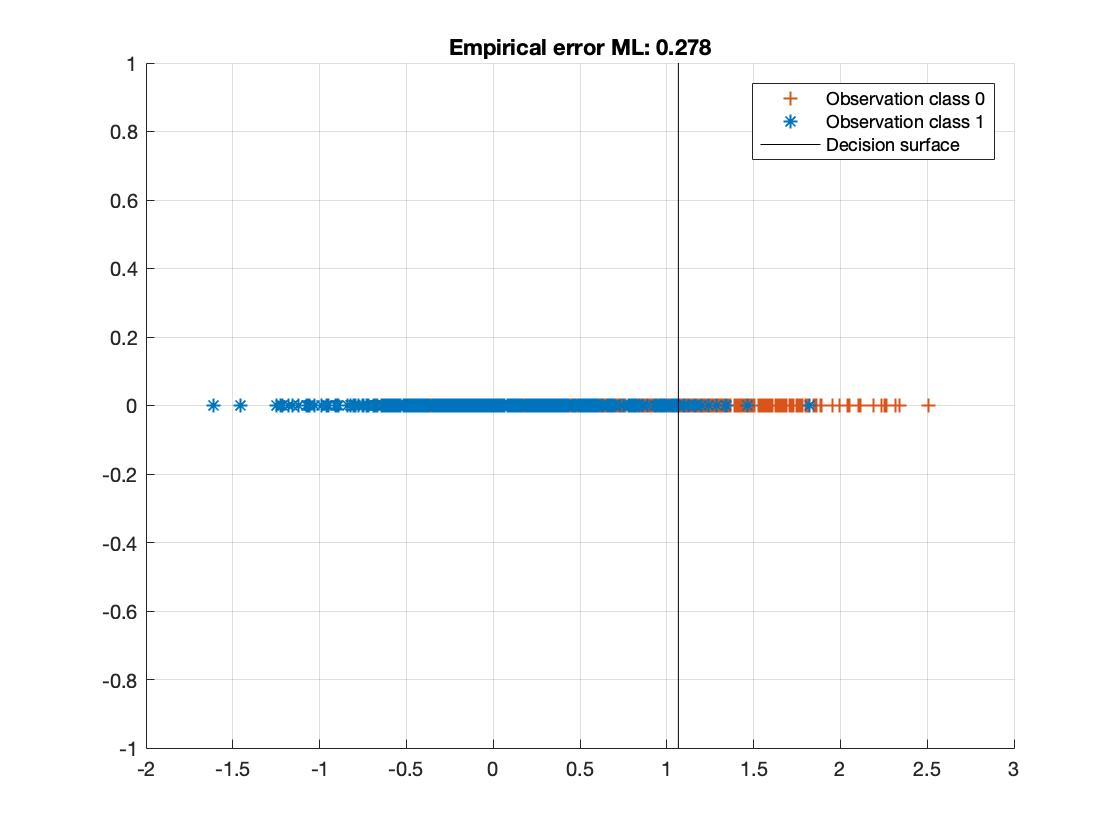}
\includegraphics[width=0.49\textwidth]{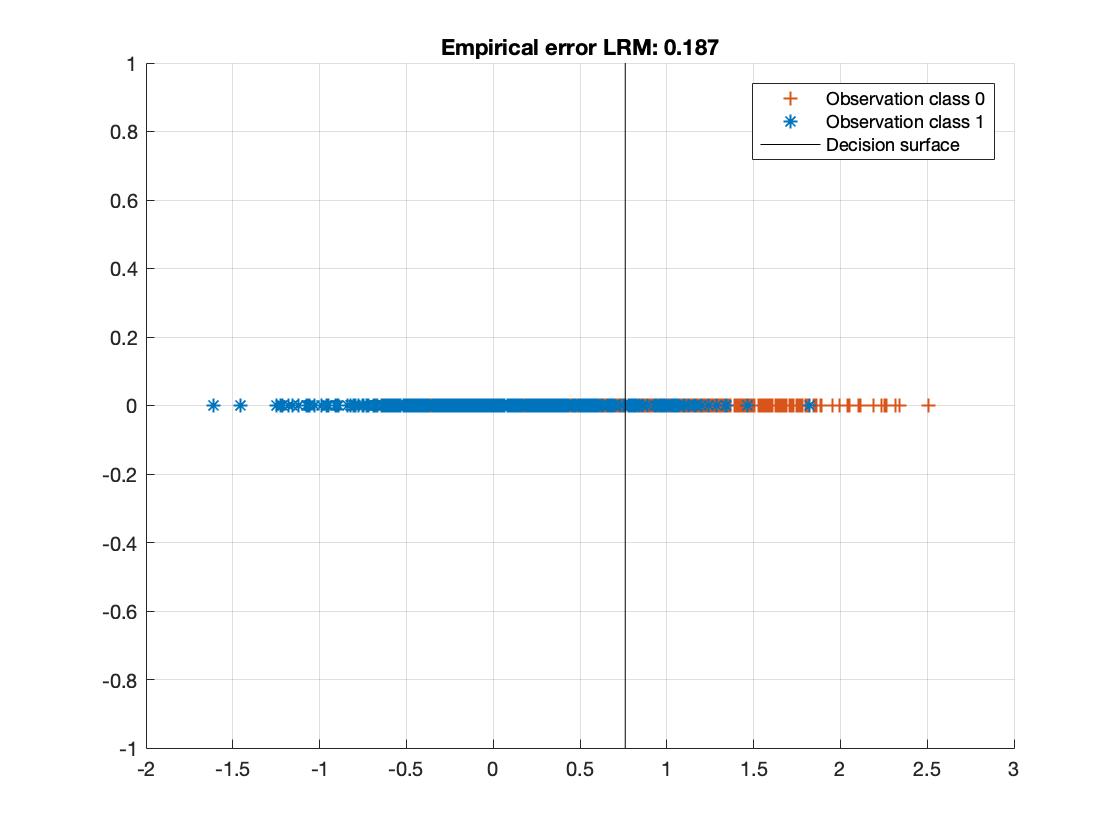}
\includegraphics[width=0.49\textwidth]{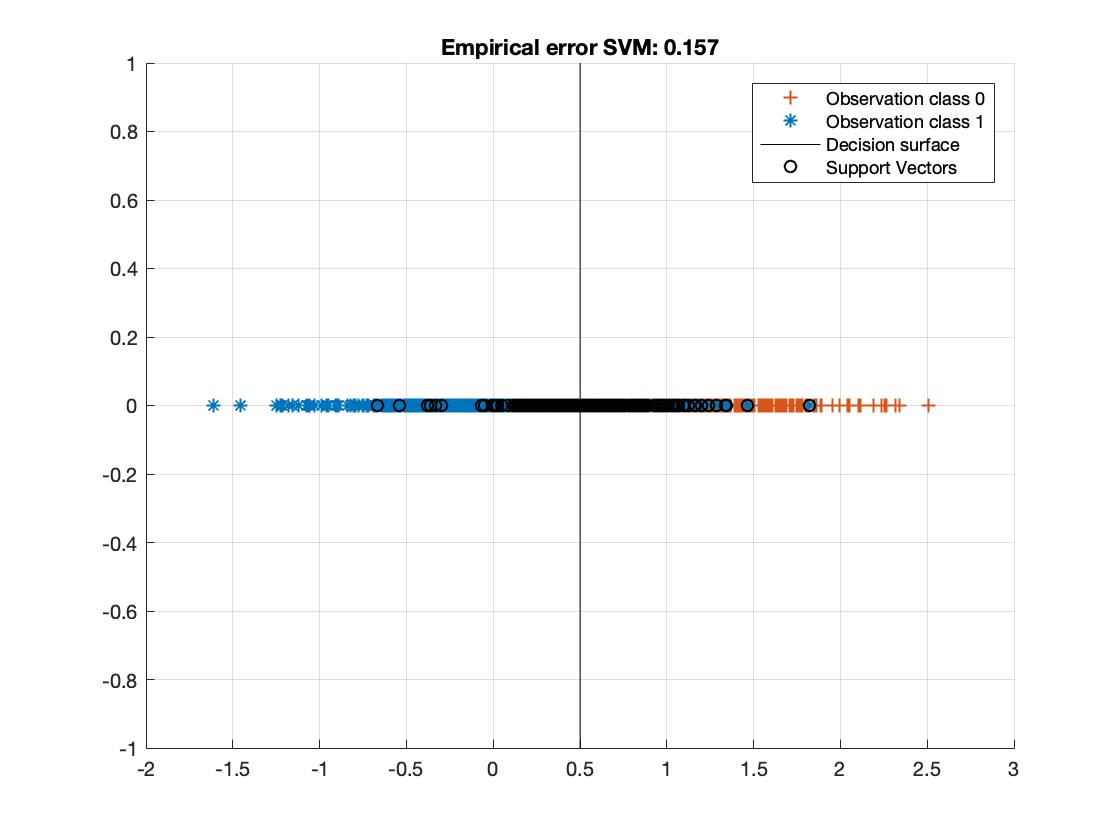}
\caption{Classification boundaries and empirical errors in GLM,  LRM and SVM (N=1000, DG1).}
\label{fig:cinco}
\end{figure*}

In connection with the previous one-sample GLM estimations, we plot the estimated parameters explaining the observation using all the methods in figure \ref{fig:tres} together with the observations they model. Note the large variability of the GLM estimation with increasing sample size. On the contrary, in figure \ref{fig:cuatro} we show the inverse problem to the above: how the methods estimate the $\mathbf{w}$ from the point of view of the label regression. In this case it is readily seen that the one sample GLM model provides a wrong estimation at different sample sizes, i.e. red curve above blue curve. As expected, the use of these parameter in the dual classification problem results in a larger empirical error as shown in \ref{fig:cinco}.

\subsection{Data generation 2}

\begin{figure*}
\centering
\includegraphics[width=\textwidth]{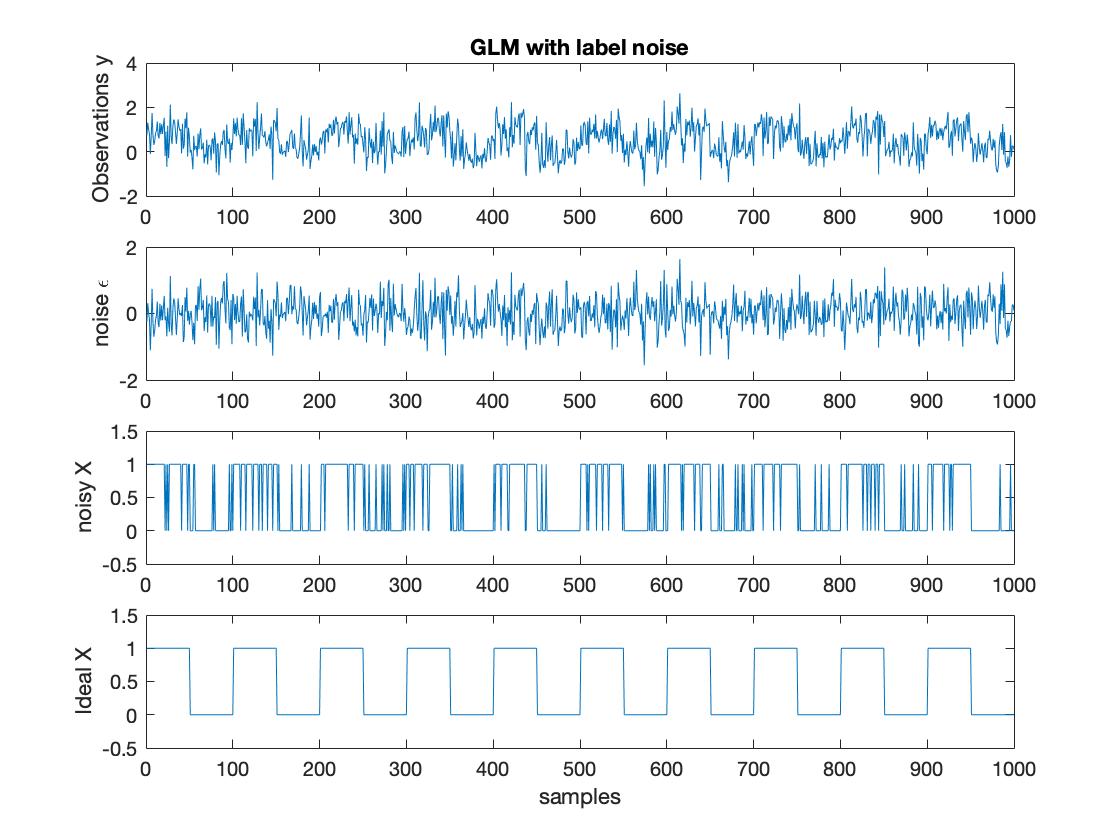}
\caption{DG 2 example}
\label{fig:unobis}
\end{figure*}

\begin{figure*}
\centering
\includegraphics[width=\textwidth]{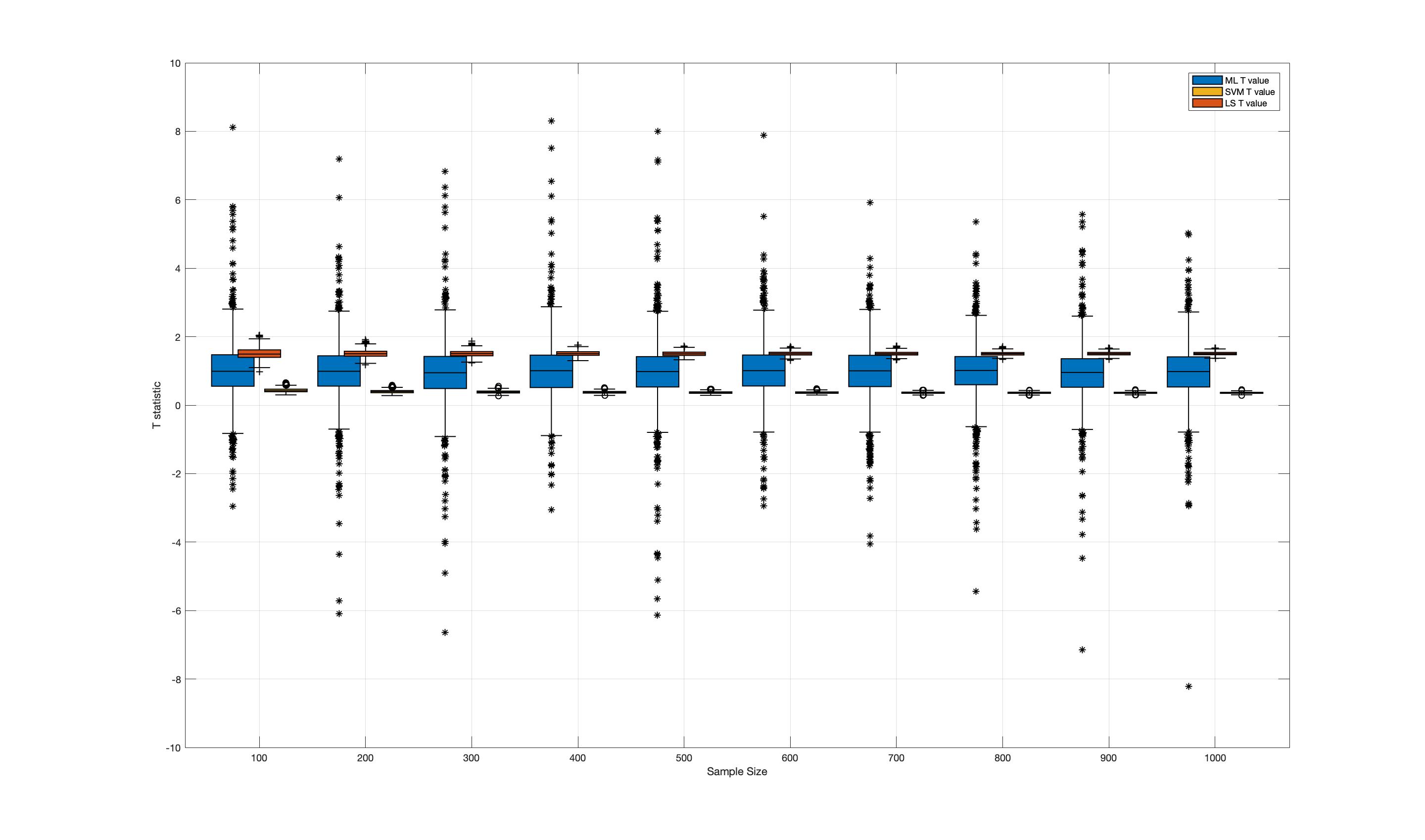}
\includegraphics[width=0.49\textwidth]{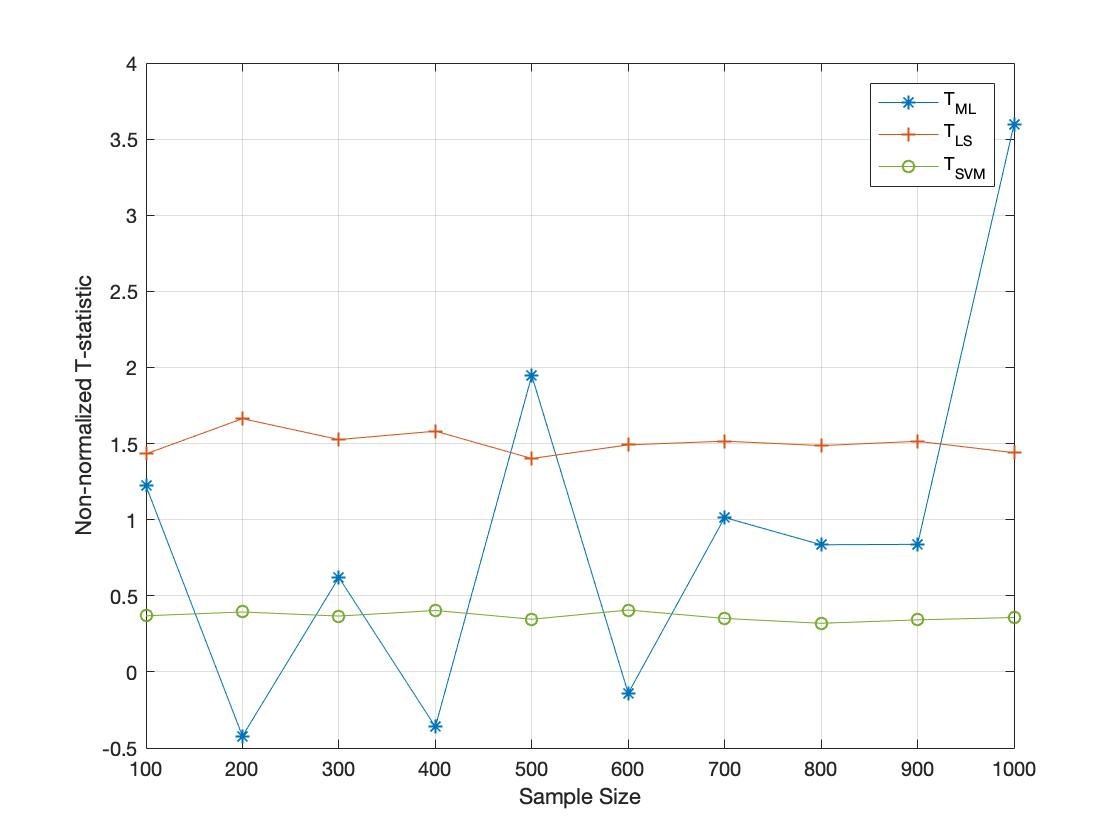}
\includegraphics[width=0.49\textwidth]{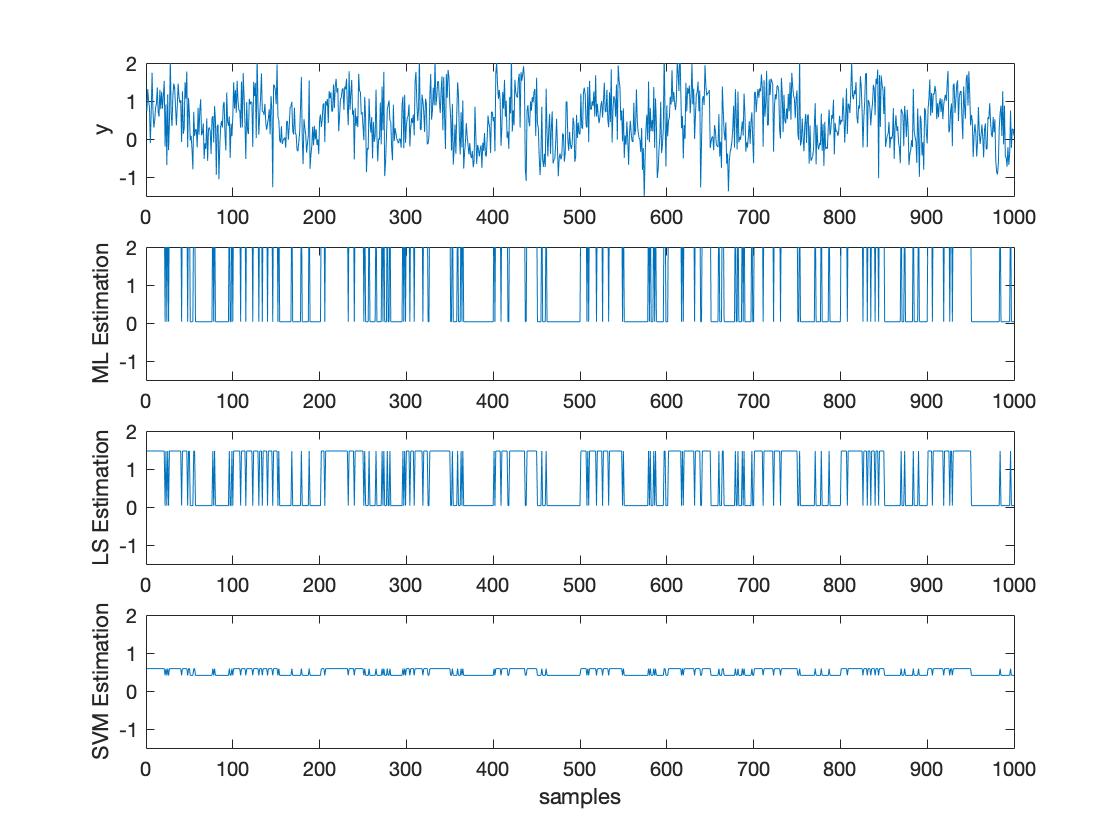}
\caption{Estimated Observations and T-statistic distribution. (DG2)   See caption in figure \ref{fig:uno}}
\label{fig:dosbis}
\end{figure*}

\begin{figure*}
\centering
\includegraphics[width=\textwidth]{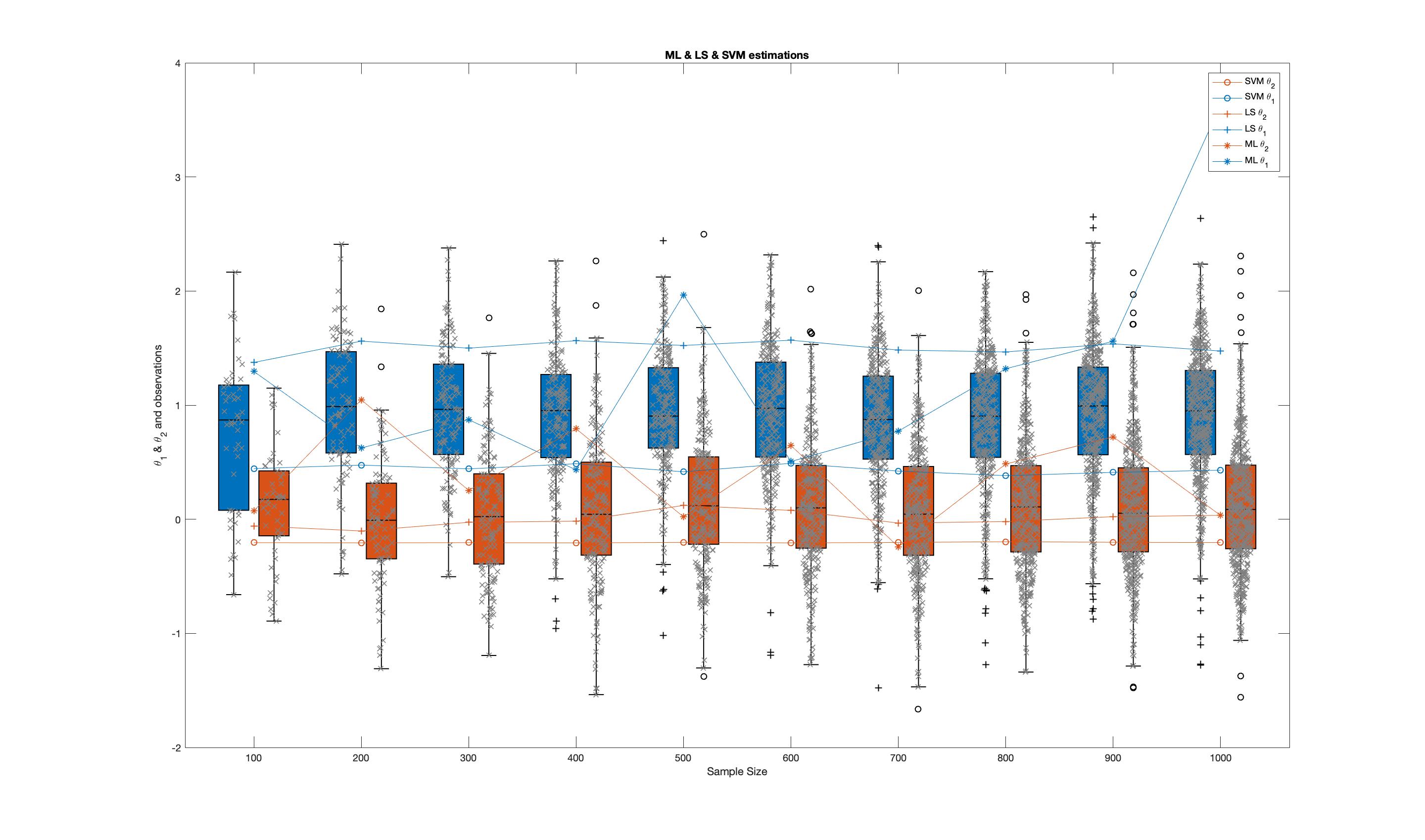}
\caption{Distribution of observations and estimations of  $\mathbf{\theta}$ for GLM,  LRM and SVM (DG2).}
\label{fig:tresbis}
\end{figure*}

\begin{figure*}
\centering
\includegraphics[width=\textwidth]{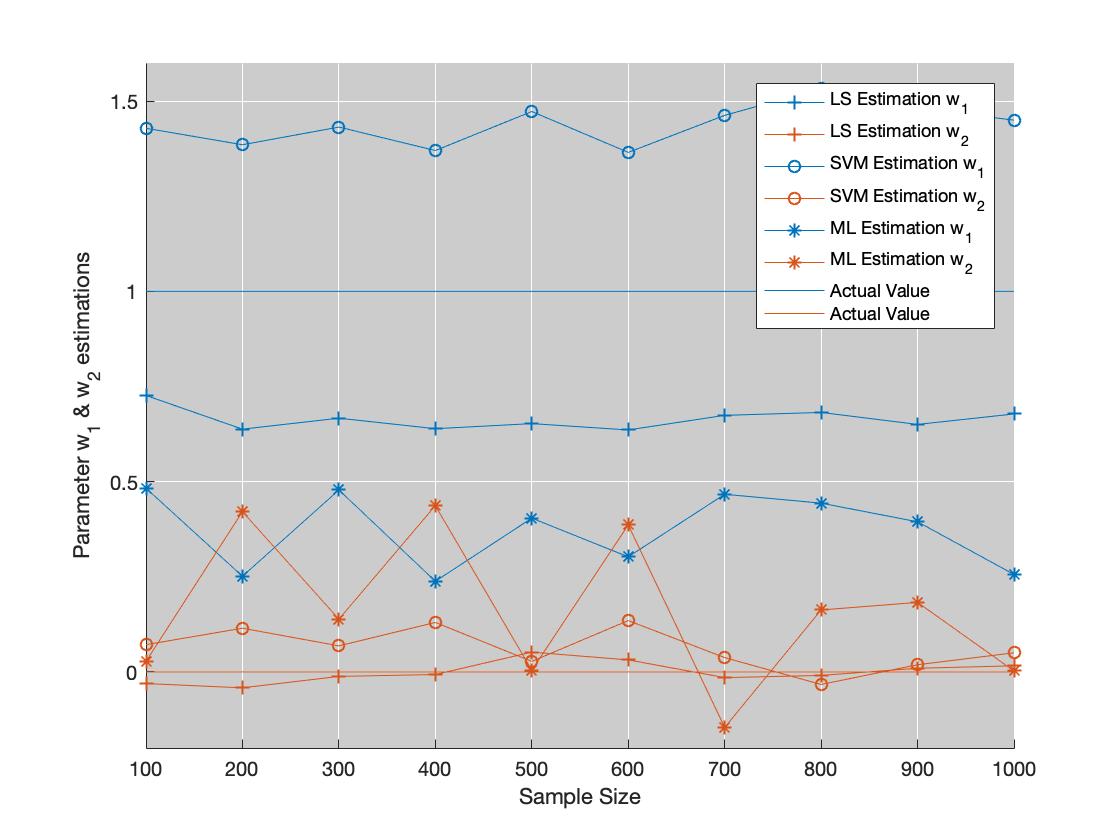}
\caption{Estimations of the parameter $\mathbf{w}$ regressing the observations with increasing sample size in DG2}
\label{fig:cuatrobis}
\end{figure*}

\begin{figure*}
\centering
\includegraphics[width=0.49\textwidth]{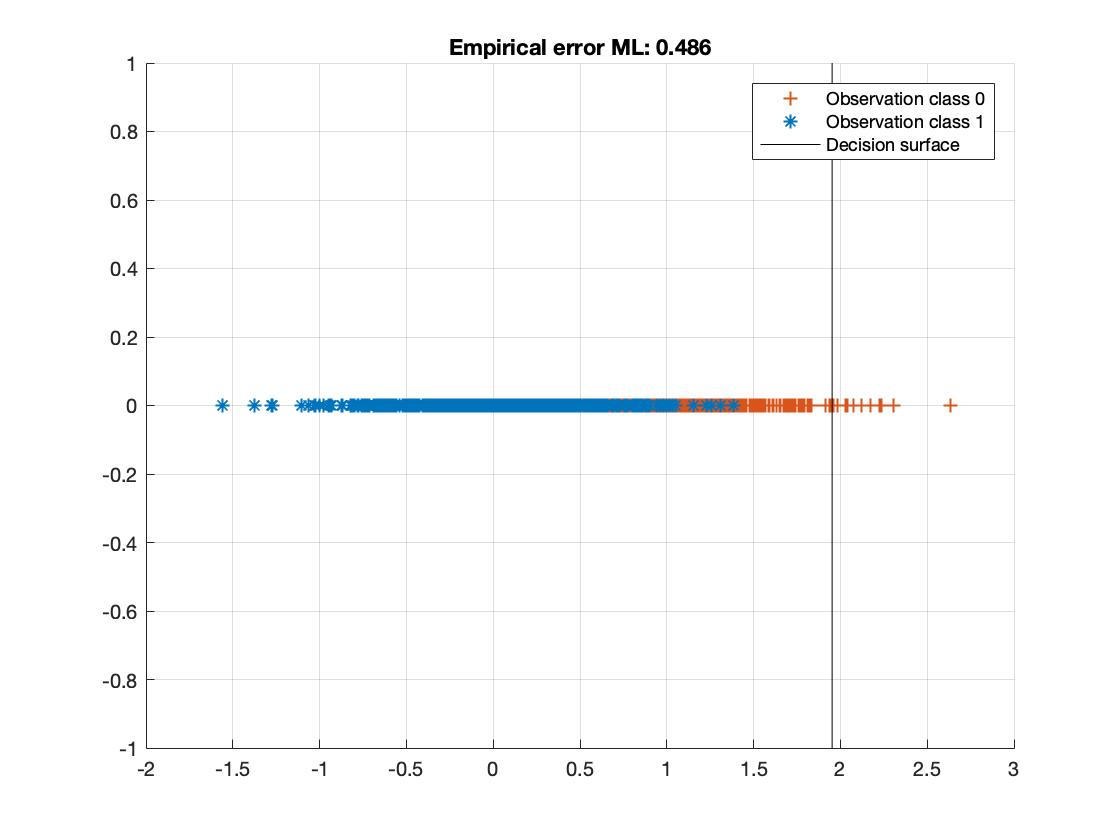}
\includegraphics[width=0.49\textwidth]{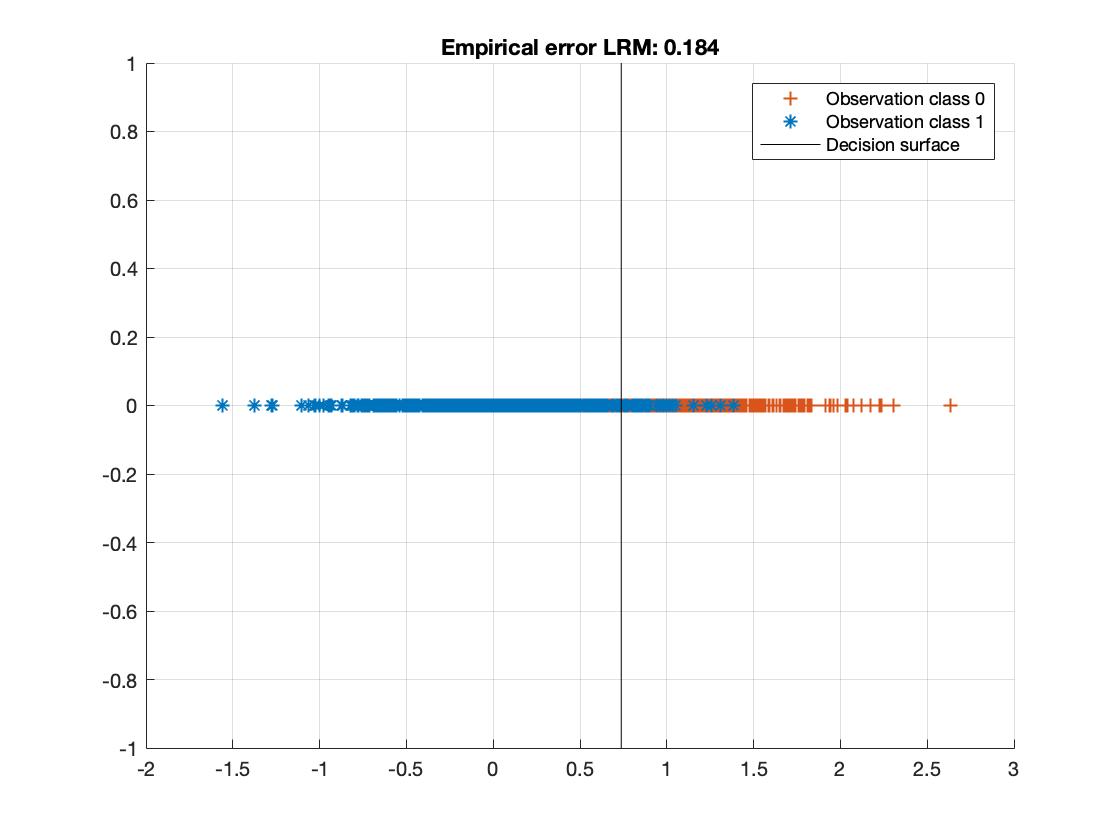}
\includegraphics[width=0.49\textwidth]{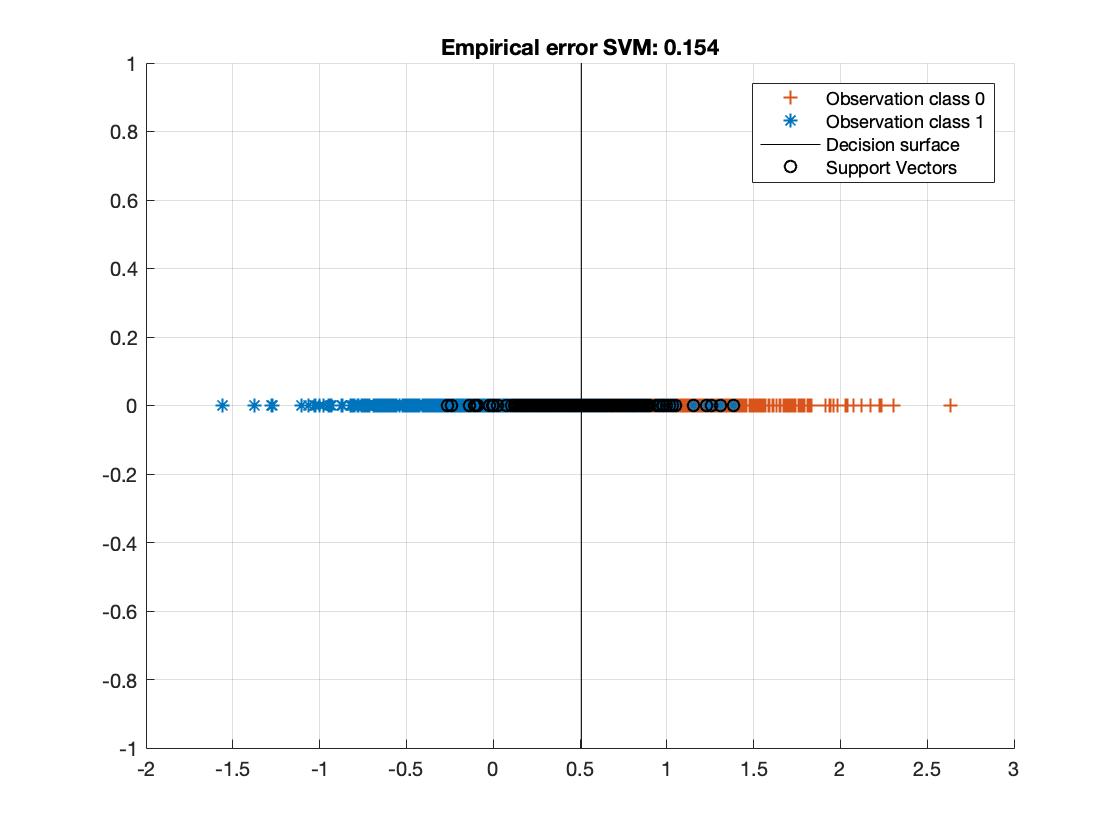}
\caption{Classification boundaries and empirical errors in GLM, LRM and SVM (DG2)}
\label{fig:cincobis}
\end{figure*}

A more realistic data generation is described in the following (DG2). The previous model is a standard procedure, where two expected values ($1$ and $0$) are added to the same Gaussian noise model. Here we propose to introduce the concept of label noise in the design matrix, instead of only in the observations. Again $\mathbf{y}=\hat{\mathbf{X}}_k+\mathbf{v}$, but the matrix of indicators $\hat{\mathbf{X}}$ is a version of the original, equal to it with probability $1-t$, with $t=0.1$, and flipped with probability $t$. This allows us to control the noise label in the design matrix, following the methodology shown in \cite{Ojala2010}.  
We repeat the same experiments (see figures \ref{fig:unobis}-\ref{fig:cincobis}) of the previous section using this realistic data generation.  Again, the observations, error and explanatory matrix (ideal and nosy version) are shown in \ref{fig:unobis}. Note how this observation model is more realistic than the previous one and corresponds to a group comparison including wrongly labeled subjects or acquisitions. From figures \ref{fig:dosbis} and \ref{fig:tresbis} the same behavior could be expected as in the previous case. However, when we analyze the inverse problem we see a huge difference in the GLM estimation of $\mathbf{w}$. The dependence of the GLM estimator on the one-point sample mean provides a fluctuating estimation about the optimum value, unlike LRM and SVM. It is also worth mentioning that, in this realistic example, the estimation of $\mathbf{w}$ using GLM provides a huge empirical error as shown in figure \ref{fig:cincobis}.

As a conclusion, the link between the two approaches is the different nature of the regression procedure.  In both domains there is an implicit classification task once the parameters, that better explain the corresponding observations, are derived. These parameters are fitted taking into account only the empirical data available (including a noise model or not). Therefore, $w_{m}$ for a given model $m$, can be used to regress the observations to obtain a novel data set on the label space (new regressed labels), which can be associated to the states (or classified) of the explanatory matrix. This classification task provides an empirical error as shown in figures \ref{fig:cinco} and \ref{fig:cincobis}.  Other methods could be used as well to obtain such parameters in a (non) linear fashion.  As an example, we compared the decision boundary obtained by LRM with SVM in figures \ref{fig:cinco} and \ref{fig:cincobis} to show the differences between methods in terms of generalization ability.

\subsection{Real data experiment: the ADNI dataset}

Data used in preparation of this paper were obtained from the Alzheimer's Disease Neuroimaging Initiative (ADNI) database (adni.loni.usc.edu). The ADNI database contains 1.5 T and 3.0 T t1w MRI scans for AD, Mild Cognitive Impairment (MCI), and cognitively NC which are acquired at multiple time points. Here we only included 1.5T sMRI corresponding to the three different groups of subjects. The original database contained more than 1000 T1-weighted MRI images, comprising 229 NC, and 188 AD,  although for the proposed study, only the first medical examination of each subject is considered, resulting in 417 gray matter (GM) images. Following the recommendation of the National Institute on Aging and the Alzheimer's Association (NIA-AA) for the use of imaging biomarkers \cite{NIA18}, we considered the group comparison NC vs. AD for establishing a clear framework for comparing statistical paradigms (SPM and $T_{CV}$),  since the MCI class is strictly based on clinical criteria, without including any other biomarker \cite{McKhann11}. Demographic data of subjects in the database is summarized in Table \ref{tab:demog}. The dataset was preprocessed using standardised neuroimaging methods and protocols implemented by the SPM software (registration in MNI space by spatial normalization and segmented to differentiate brain tissues, e.g.  GM \cite{Friston95}. 

\begin{figure*}
\centering
\includegraphics[width=\textwidth]{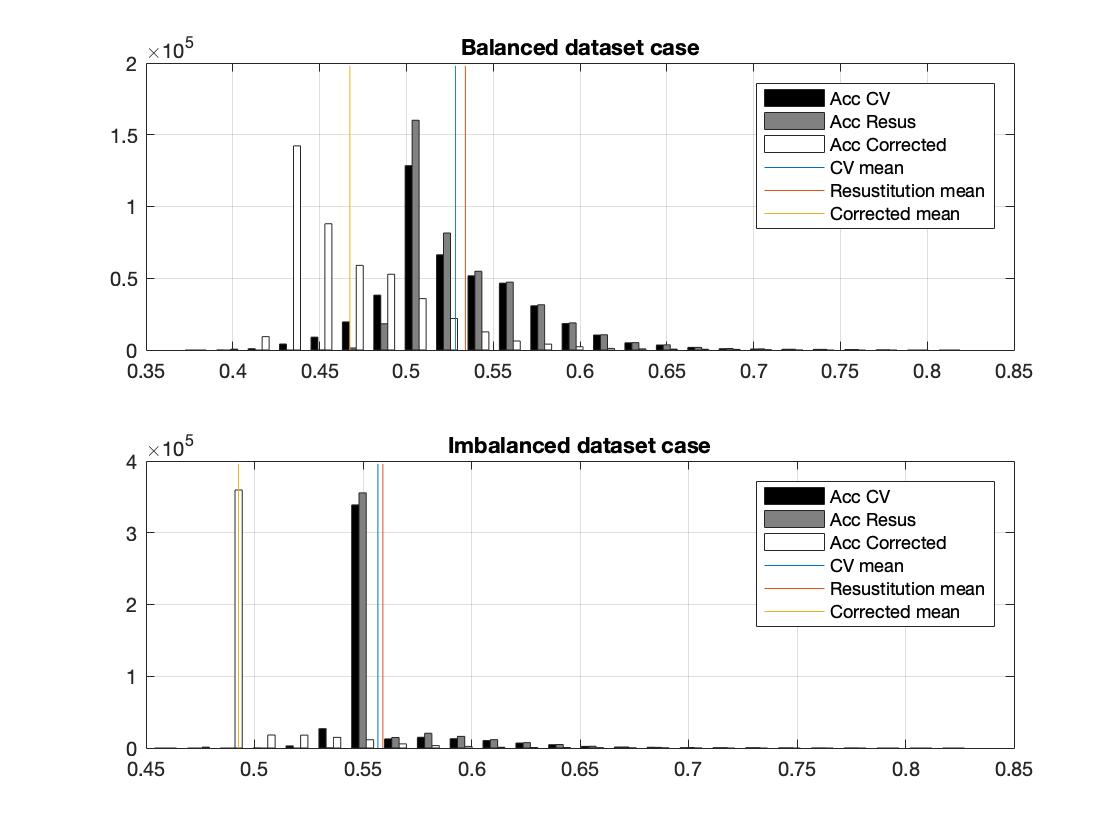}
\includegraphics[width=0.49\textwidth]{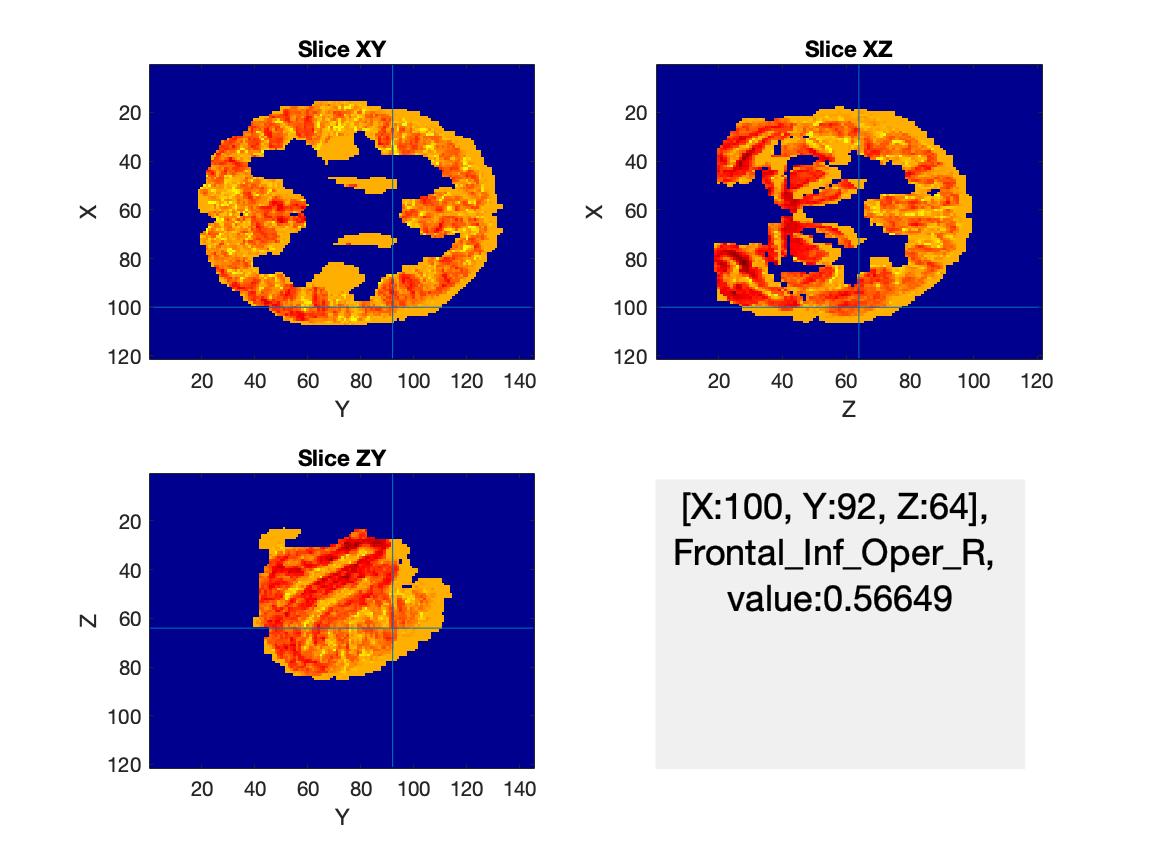}
\includegraphics[width=0.49\textwidth]{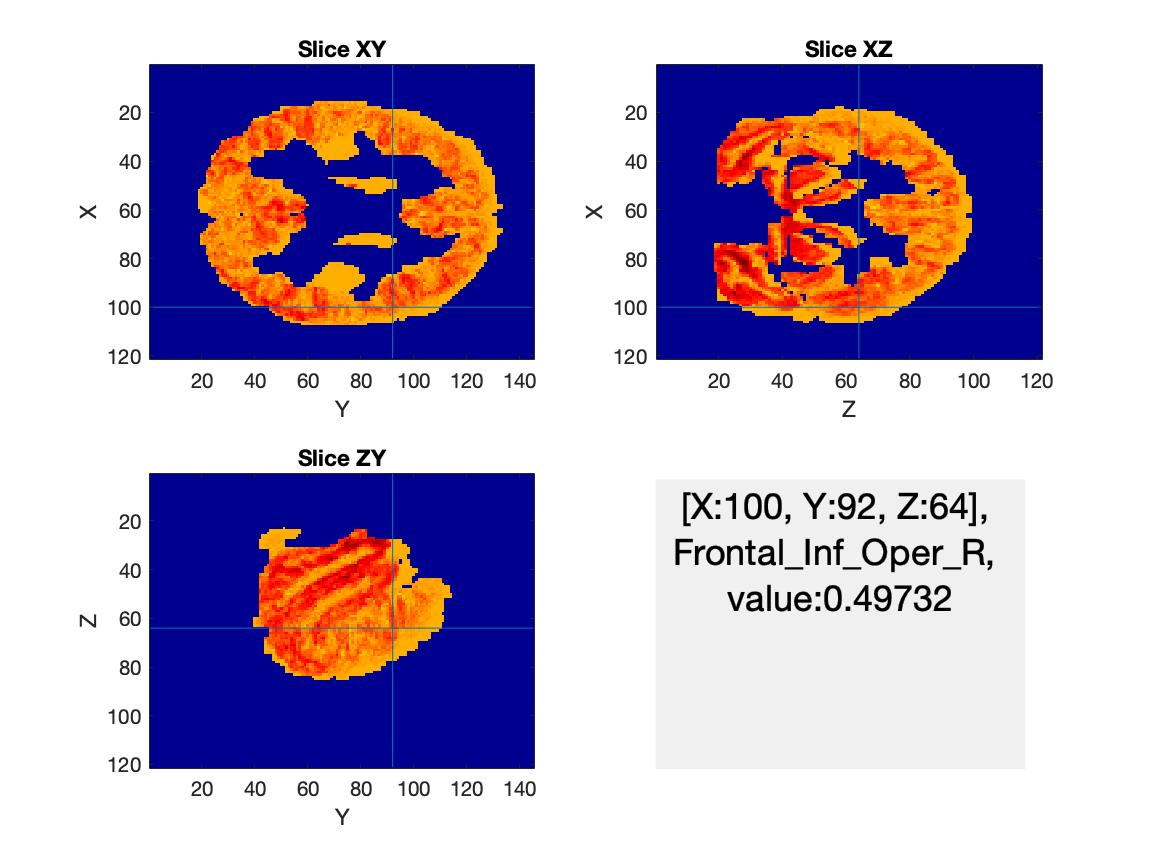}
\caption{Up: Distribution of voxelwise accuracies of the real dataset using $K=10$-fold,  resubstitution and concentration inequalities \cite{Gorriz2021}.  Down 3D distribution of the accuracies using CV and Corrected Accuracy.}
\label{fig:seis}
\end{figure*}

Following the aforementioned methods, we fit the set of parameters using linear SVM and evaluate the $T_{CV}$ statistic on the original set (see figure \ref{fig:seis}). As shown from this figure the resubstitution estimate provides a more optimistic value in the Acc distribution than the $K$-fold based estimate. Note that this analysis is independent of the selected fold as we are performing $\sim 10^6$ folds, one per voxel. However, both are optimistic since the mean of the distribution is not clearly distributed around $0.5$ (it is already shifted to the right, beyond the effect due to real significant regions). The effect is even larger when the dataset is slightly imbalanced, the case of over-powered datasets, as shown in the bottom of the latter figure (using 228vs188).  On the contrary, note how the correction based on the bound derived in \cite{Gorriz19} clearly shifts the $A_{cc}$ obtained by resubstitution to the left, resulting in a better (conservative) estimation of the statistic in the whole volume.  

Based on the $T_{CV}$ and $T_{Res}$ values from the original dataset, and the ones obtained using a permutation analysis ($O=1000$) for a selection of structures,  e.g. hippocampus, we can compare the SPM with the previous inference approaches, as described in section \ref{sec:inference}.  Note that in this paper the huge amount of voxels contained within an image limits the permutation analysis in this sense to some specific structures. Results on the hipocampus are depicted in figure \ref{fig:siete}. The permutation analysis reveals how the power of the $T_{CV}$ approach is affected in this featured region, where a real effect might be found in almost the whole structure. The statistical power of the $T_{Res}$ is preserved through the permutation procedure ($2058$ detected voxels vs $1024$ voxels as shown in the same figure). It is also worth mentioning the CDF of the errors derived in the specific region and the distribution of the p-values within it. Recall that the dataset include advanced AD subjects thus the selected structure should be clearly affected by the disease.

To preliminary extend the analysis to the whole volume we approximately simulate the null distribution outside this featured region in two steps. First, we compute the set of p-values in the hippocampus (around $2\cdot 10^3$ voxels) following equation \ref{eq:16} and determine the T threshold $T_{th}$ that approximately provides the significance level, e.g. $0.05$. Then, assuming that for any $T<T_{th}$ the probability of observation is $p-value<0.05$, we threshold the rest of the image to obtain the significant voxels showing an effect. This approach clearly needs the multiple-comparison correction as several dependent or independent statistical tests are being performed simultaneously at the given significance level. Therefore, we decrease the significance level down to $\alpha=0.001$ to avoid the presence of false positives in permutation analyses and then compare with SPM in the whole volume using the aforementioned configurations. In figure \ref{fig:siete} we show the detection ability together with the control of type I error in the $T_{Res}$ approach (map in red font). Note how the permutation test affects the detection ability of the classical CV approach (map in green font) and how the uncorrected voxelwise SPM approaches (in blue font ) tends to inflate false positives.

\begin{table}[htbp]
   \centering
   \caption{Demographics details of the ADNI dataset, with group means with their standard deviation}
   \label{tab:demog}
   \begin{tabular}{lccccc}
     \hline
     & Status & Number &	Age	& Gender (M/F) &	MMSE\\ \midrule
MRI ADNI& NC	        &   229    &	  75.97$\pm$5.0	    &   119/110	  & 29.00$\pm$1.0 \\
& AD	        &  188	   &     75.36$\pm$7.5 	&     99/89	  & 23.28$\pm$2.0  \\ \bottomrule
\end{tabular}
\end{table}

\begin{figure*}
\centering
\includegraphics[width=0.8\textwidth]{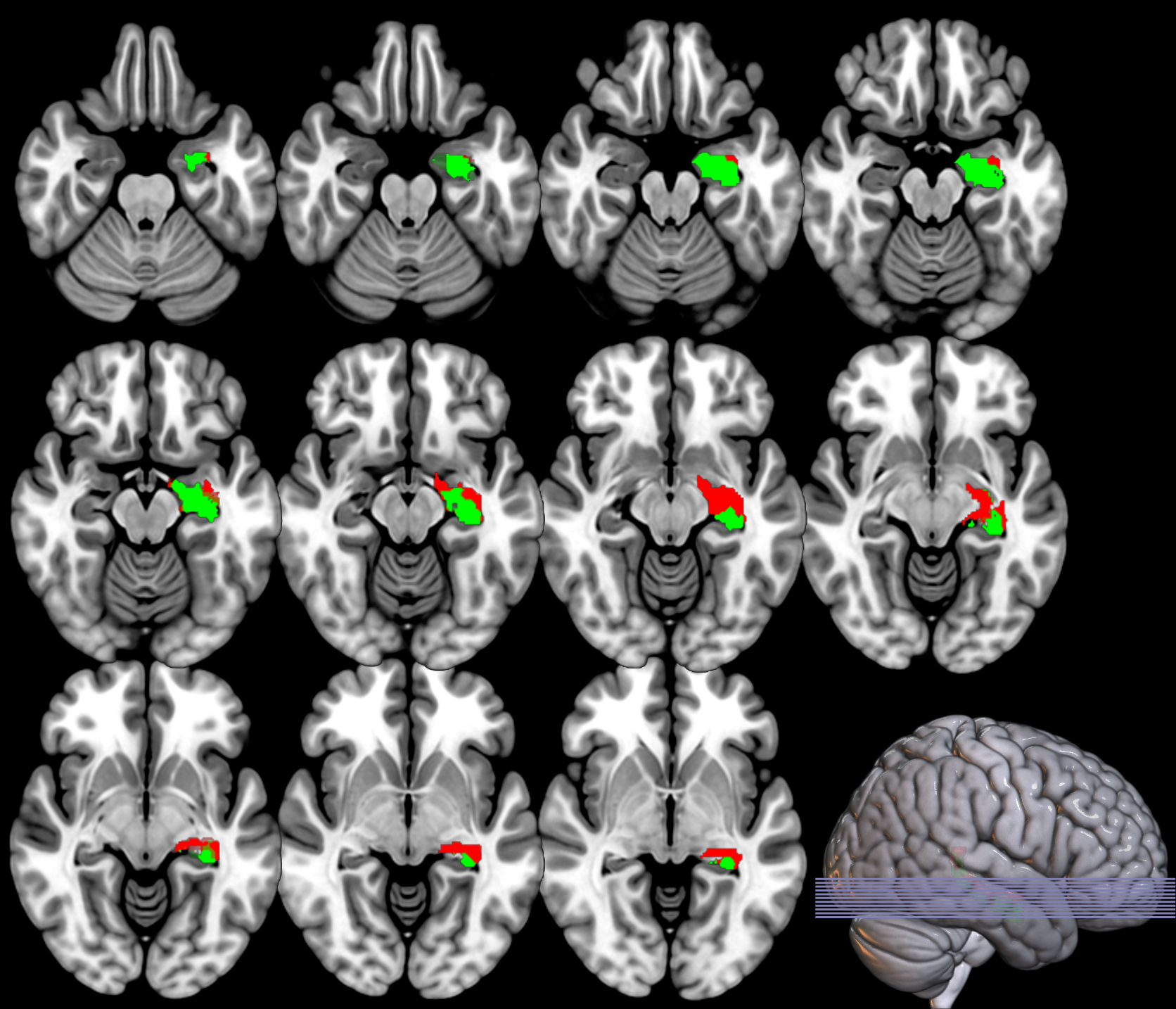}
\includegraphics[width=\textwidth]{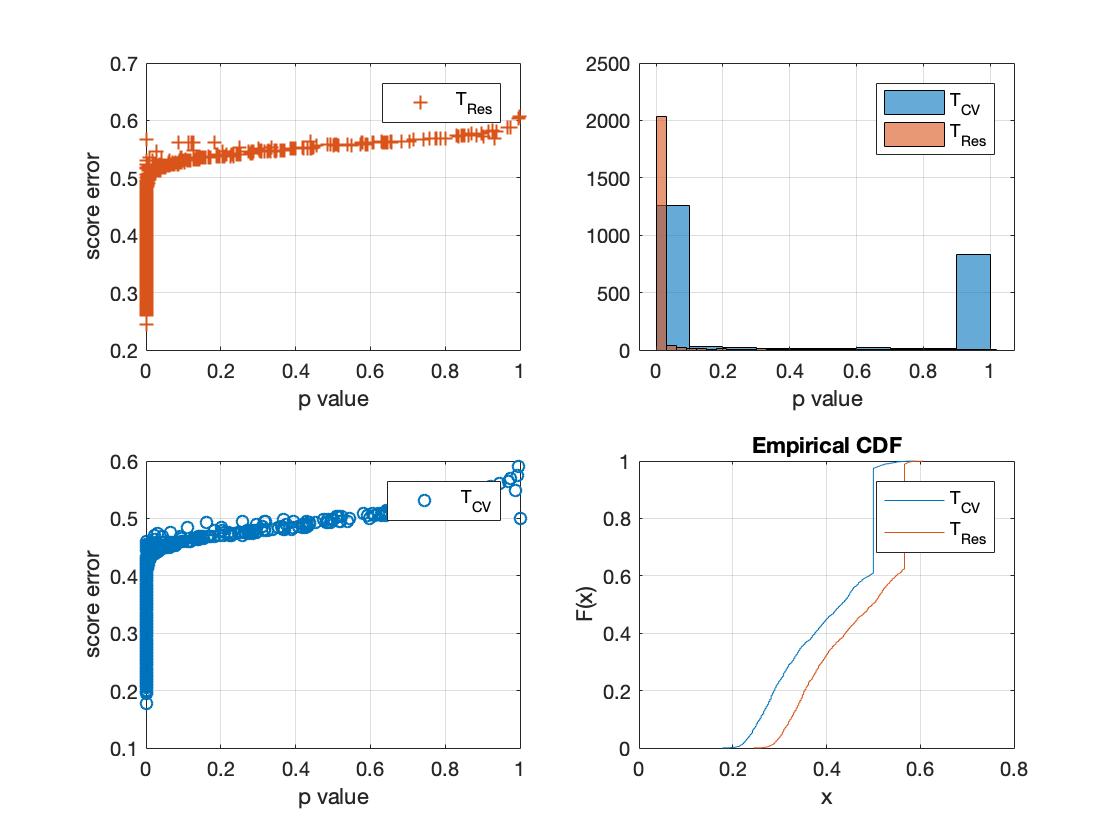}
\caption{Permutation analysis on the hippocampus. Note that $O=1000$ and the upper bound \cite{Gorriz19} was obtained with a probability at least equal to $0.05$}
\label{fig:siete}
\end{figure*}

\begin{figure*}
\centering
\includegraphics[width=0.49\textwidth]{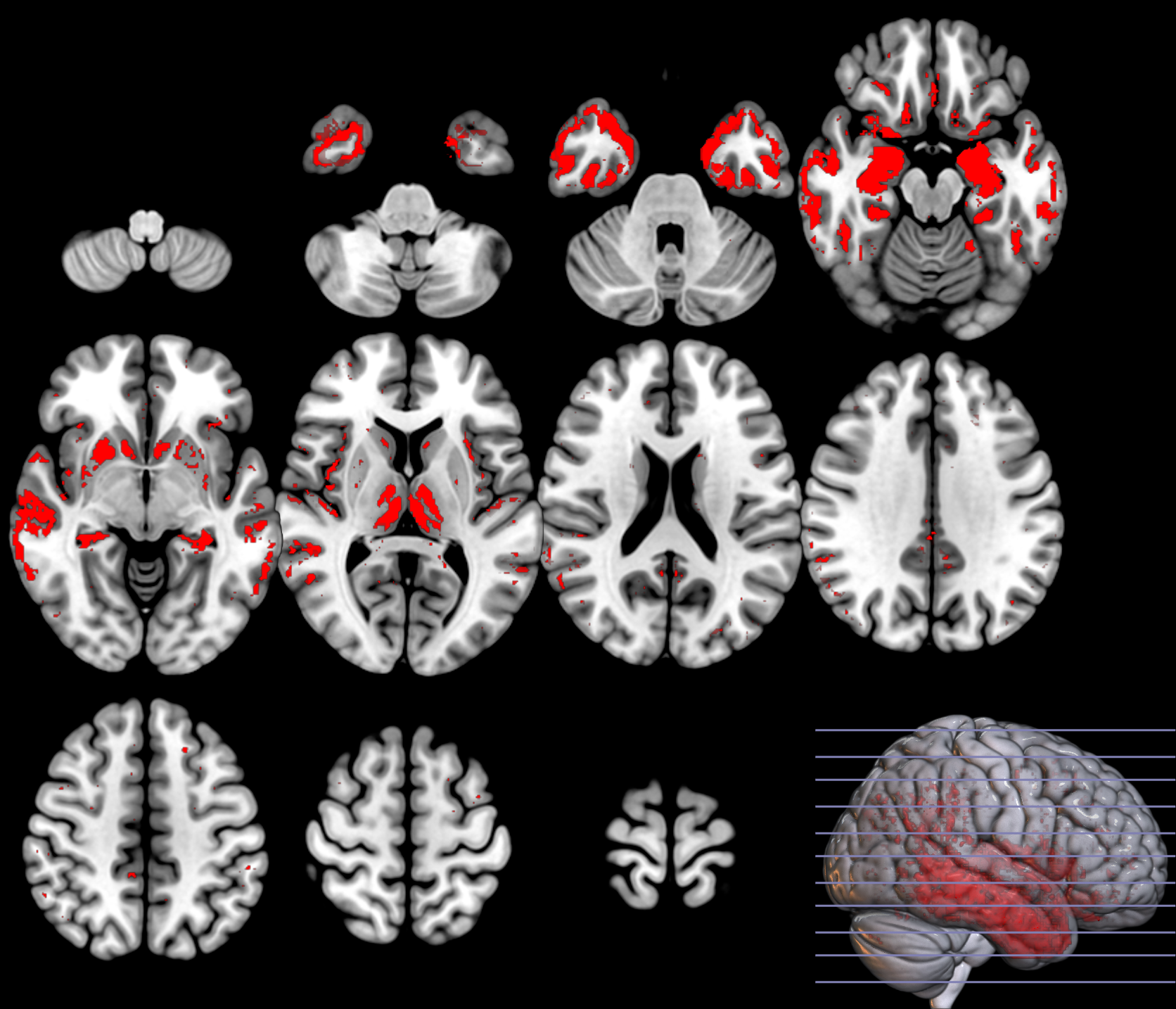}
\includegraphics[width=0.49\textwidth]{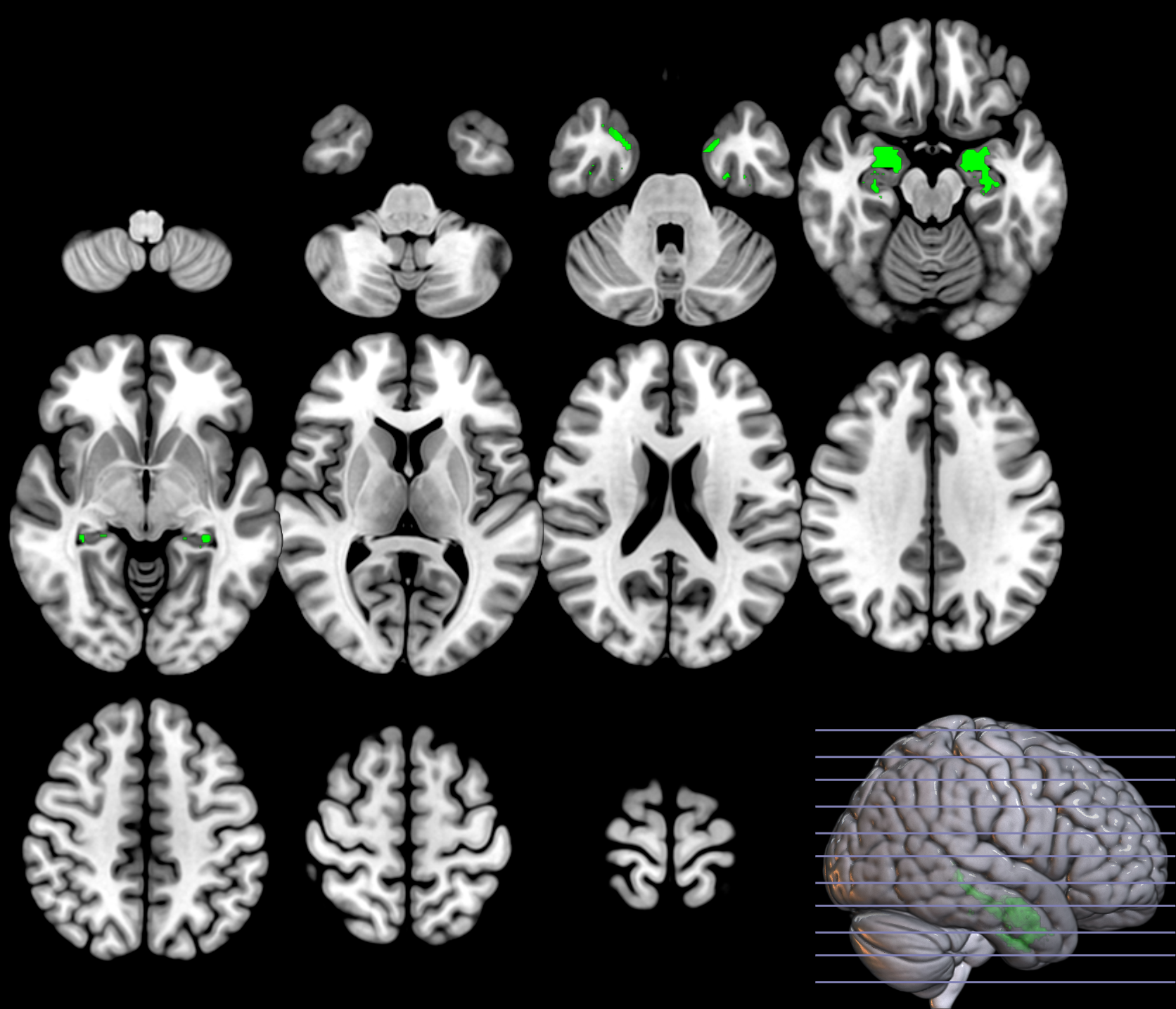}
\includegraphics[width=0.3\textwidth]{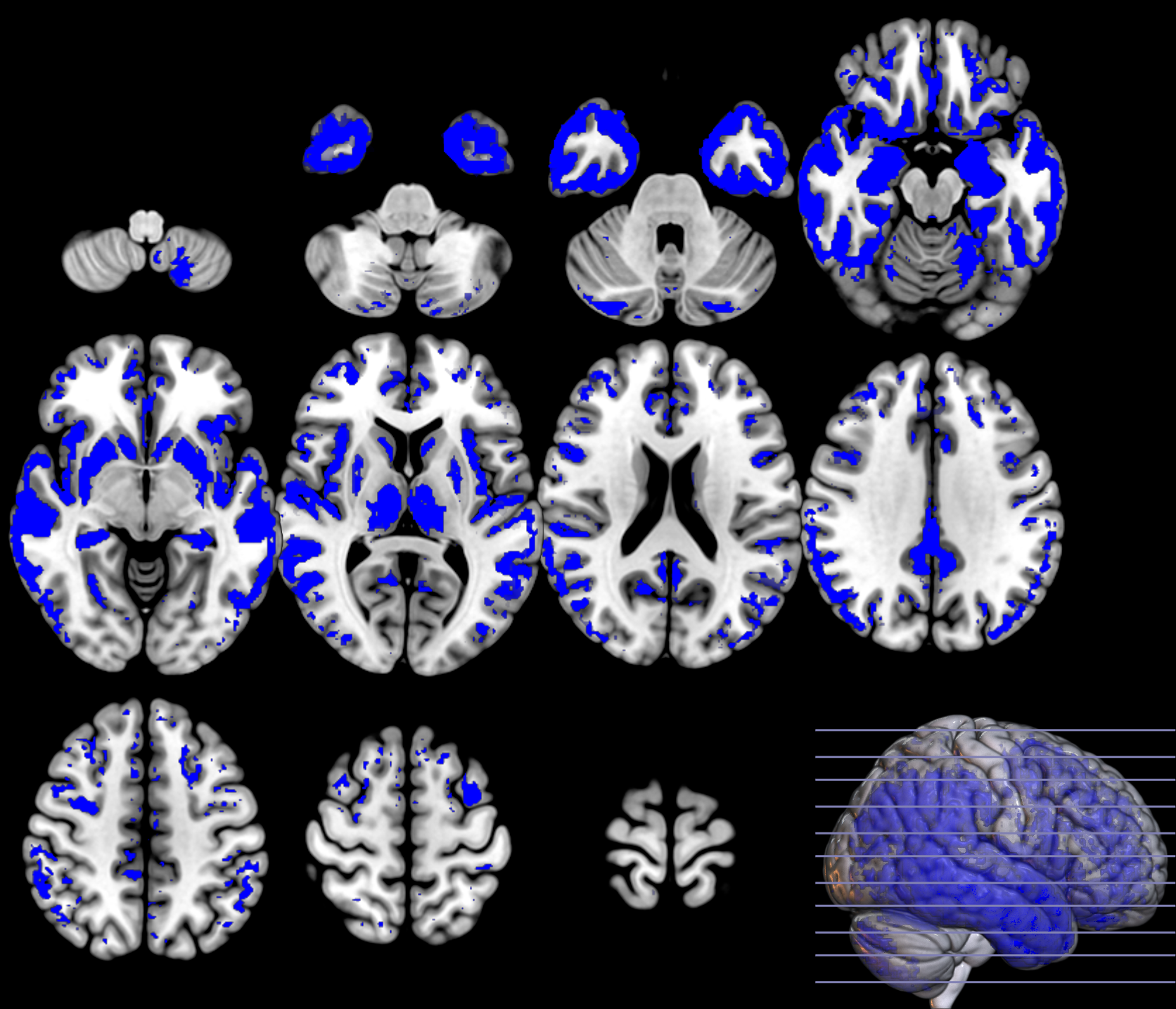}
\includegraphics[width=0.3\textwidth]{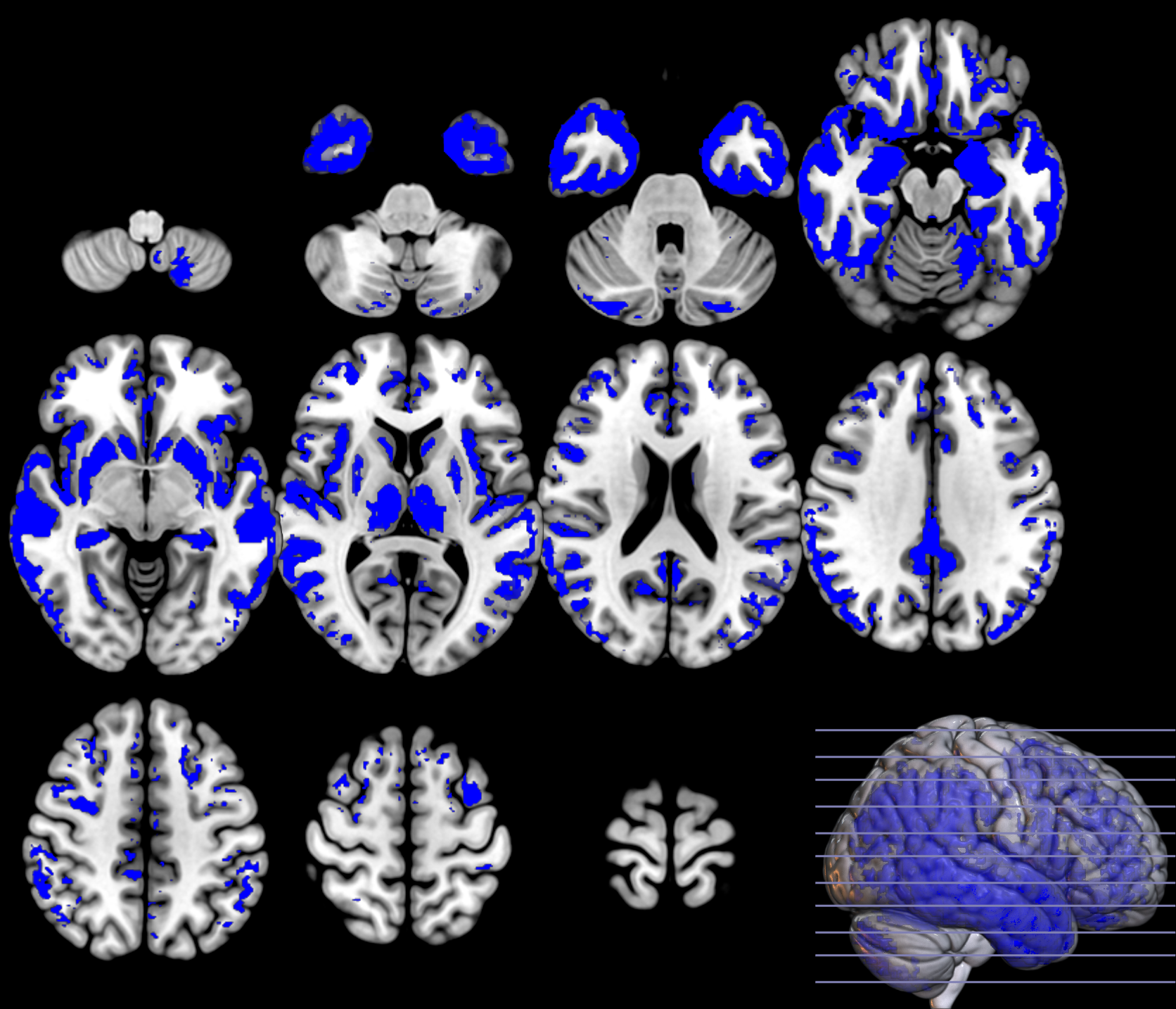}
\includegraphics[width=0.3\textwidth]{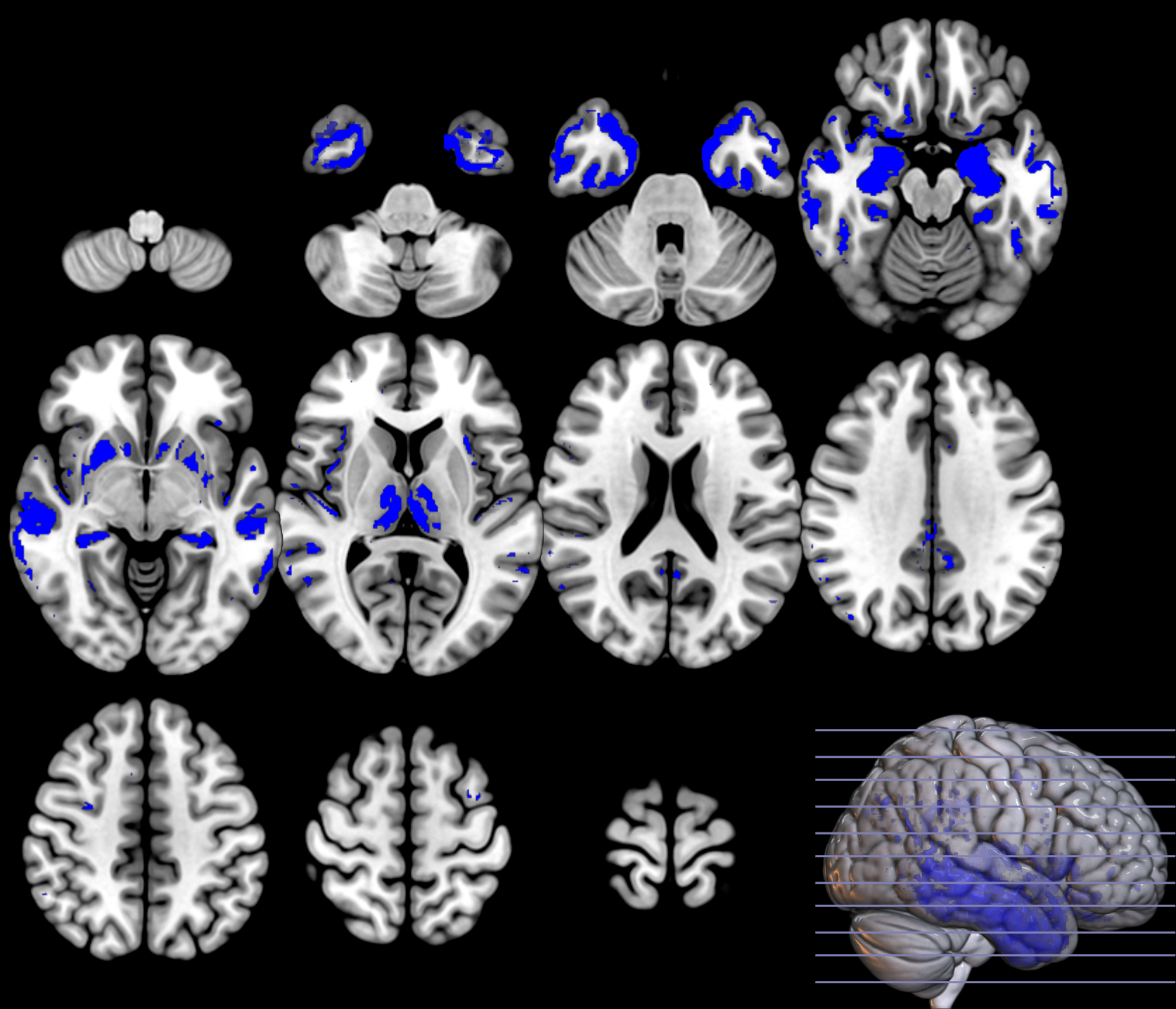}
\caption{Parametric and non-parametric statistical maps. Note the trade-off in etection and control of the FWE of the $T_{res}$ approach, compared with $T_{CV}$ and he three SPM configurations}
\label{fig:siete}
\end{figure*}

\section{Discussion}

In the context of classification for statistical inference,  researchers  mainly follow two strategies: i) they perform K-fold and assess the $A_{cc}$ in several averaged folds or ii) they propose some kind of cross-validation based statistic (P-test) using an estimation of the actual error of the classifier on new set of samples (equation \ref{eq:8}). In both cases, if this residual square (error) is small, i.e. a good classification) is achieved,  then it constitutes evidence against the null hypothesis. In the second approach to simulate the null distribution they employ a technique, as shown in section \ref{sec:inference}, that is also used in the frequentist inference, the permutation test. A set of permutations $\pi_p$ for $p=1,\ldots,O$ is generated and then applied to the dataset, using the same observations $\mathbf{y}$ and permuted constraints $x_{\pi_p}$. They estimate the parameters $w_{pi_p}$ and compute a set of residuals for all the permutations. The p-value is computed by dividing the number of times we randomly obtain a \emph{residual score} less than the one we obtain with the original value over the number of permutations, i.e. $p-value=p(RS_{\pi}<RS)$. This methodology is called CV-$P$ tests \cite{Reiss15} where LRM might be replaced by SVM or other predictive algorithms.

Several limitations are found using only LRM for estimating the posterior probability above mentioned. Linear regression is only operative in binary classification, e.g the regression could be negative or even greater than zero \cite{Hastie2001}. Indeed, as shown in the example, in this case there is a strong correspondence between GLM and LRM for a single level analysis in group comparisons. Thus complex classifiers and other loss functions are needed for relieving bad estimations on the set of parameters.  Beyond that, the selected predictive algorithms build their P-tests on the CV strategy, that could be a biased estimator of the actual error in heterogeneous datasets, such as the ones used in neuroimaging \cite{Varoquaux18}.

On the other hand, classical and Bayesian Inferences depends on the specified models when proposing the T statistic and fitting parameters of the GLM. This is partly solved again by the use of permutation analysis in the estimation of the null distribution,  but what about the T statistic definition? It is also described in terms of the error covariance matrix, which must be estimated on empirical data in limited sample sizes. In the toy example shown above we assumed in the formulation of the GLM a known covariance matrix. Despite that, the T-statistic following from the best guess was fluctuating around the ideal value and resulted in low classification rates in the dual problem. How is actually performed frequestist analysis or Bayesian analysis (equivalent to the latter in the last level) in a real scenario? Again, there are model selection and parameter fitting stages to achieve that, nonetheless in complex scenarios with a limited sample size, heuristics are the common solutions \cite{Woolrich2009}. Indeed, in the high dimensional case or under the assumption of complex models, the performance and operation of the latter approaches is arguable \cite{Reiss15}. They hardly estimate parameters, the computation is costly and tend to use heuristics for solving such issues, e.g. in the FSL tools based on Bayesian inference, such as BET (Brain extraction tool), TBSS (tract-based spatial statistics), FLIRT (FMRIB's linear image registration tool), PRELUDE/FUGUE (phase unwarping and MRI unwarping), MELODIC ICA, the use of heuristics is a common practice and the estimation of the full posterior distribution of model parameters is biased.

As a conclusion, limited samples sizes and the selection/estimation of any specific model is still an issue in neuroimaging, further when the model and the interaction between model parameters become too complex for an accurate posterior probability estimation, or a feasible numerical computation of the Bayes rule. Given the connection between the two observation models, i.e. GLM and LRM,  in this paper we propose the use of an agnostic theory about the estimation of dependencies and established in the pattern classification problem with limited amounts of data, to achieve statistical inference \cite{Vapnik82,Haussler92}.

\section{Conclusions}

In this paper we propose the use of permutation tests and agnostic theory in the set of regressed outputs by the definition of the residual score or $A_{cc}$ based test. We employ permutation tests and a better estimation of actual error based on concentration inequalities to provide a trade-off between the Type I error and the statistical power. Some previous results demonstrate the ability of such estimator to provide maps of significance \cite{Gorriz2021} where a random simulation on controls resulted in a nominal rate of false positives.

As a conclusion, we see the equivalence in the estimation in the observation and the (explanatory) label domains, thus any test performed on the label space using an $A_{cc}$-based test is similar to the ones used in neuroimaging in the last decade. Moreover,  prevalence (scores in equations \ref{eq:15} and \ref{eq:17}) is a valid measure for statistical inference without using any model at the very first assumptions. Our approach tries to compute this score using all the database available, instead of splitting it into folds, and with the resulting set of accuracies, we estimate the real one based on the upper bounds (instead of using K-fold strategy) with probability at least $1-\alpha$. Then,  some permutation analysis is derived using this measure to simulate the distribution of the null hypothesis and finally a test can be formulated in a classic statistical sense.

\section*{Acknowledgments}
This work was partly supported by the MINECO/ FEDER under the RTI2018-098913-B100 CV20-45250 and A-TIC-080-UGR18 projects and by the Ministerio de Universidades under the FPU Predoctoral Grant FPU 18/04902. 

\section*{Appendix}

\subsection*{Proof of the connection:}
The derivation of the connection between $\mathbf{\theta}$ and $\mathbf{w}$ is shown in the following assuming non-singular matrices when needed. Given the optimum solution for the matrix of parameters $\hat{\mathbf{w}}=(\mathbf{y}^T\mathbf{y})^{-1}\mathbf{y}^T\mathbf{X}$, from the LRM in equation \ref{eq:7} we readily see that the observation can be written as:
\begin{equation}\label{eq:21}
\mathbf{y}=\mathbf{X}\hat{\mathbf{w}}^T(\hat{\mathbf{w}}\hat{\mathbf{w}}^T)^{-1}
\end{equation}
The transpose of the parameter matrix $\hat{\mathbf{w}}$ can be expressed as:
\begin{equation}\label{eq:22}
\hat{\mathbf{w}}^T=(\mathbf{y}^T\mathbf{X})^T(\mathbf{y}^T\mathbf{y})^{-T}=(\mathbf{y}^T\mathbf{y})^{-1}\mathbf{X}^T\mathbf{y}
\end{equation}
then, equation \ref{eq:19} transforms into:
\begin{equation}\label{eq:23}
\mathbf{y}=\mathbf{X}\hat{\mathbf{w}}^T((\mathbf{y}^T\mathbf{y})^{-1}\mathbf{y}^T\mathbf{X}(\mathbf{y}^T\mathbf{y})^{-1}\mathbf{X}^T\mathbf{y})^{-1}= (\mathbf{y}^T\mathbf{y})^{2}(\mathbf{y}^T\mathbf{X}\mathbf{X}^T\mathbf{y})^{-1} \mathbf{X}\hat{\mathbf{w}}^T
\end{equation}
where the leading terms in parenthesis are scalars. The first term is the squared power of the observation $(\mathbf{y}^T\mathbf{y})^{2}=(\sum_{i=1}^Ny_i^2)^2$ and the second, given that $\mathbf{X}$ is an indicator matrix and:
\begin{equation}\label{eq:24}
\mathbf{y}^T\mathbf{X}=\left(\sum_{i\in C_1}y_{i1},  \sum_{i\in C_2}y_{i2},\ldots,\sum_{i\in C_M}y_{iM}\right)=(\mathbf{X}^T\mathbf{y})^T
\end{equation}
, where $y_{im}$ denotes the observation $i$ in class $m$, can be expressed as the denominator in equation \ref{eq:13}.

\subsection*{Binary Support vector Machines:}
In general, SVM separate binary labeled training data $(\mathbf{y},x)\in \mathbb{R}^P\times\pm 1$ by the hyperplane $f:\mathbb{R}^P\rightarrow \pm 1$:
\begin{equation}\label{eq:1ap}
f(\mathbf{y})=\mathbf{w}^T\mathbf{y}+w_0=(\mathbf{w}^T w_0)(\mathbf{y}^T 1)^T
\end{equation}
where $\mathbf{w}$ is known as the weight vector and $w_0$ as the threshold. This hyperplane is obtained in such a way \cite{Burges98} that is maximally distant from the two classes, i.e. the maximal margin hyperplane.

In our case, the binary labels $x$ are coding each row of the explanatory matrix $\mathbf{X]}$, e.g. $1\rightarrow (1,0)$ and $-1\rightarrow (0,1)$, and $P=1$, thus our function $f:\mathbb{R}\rightarrow \{(0 1), (1 0)\}$ that can be easily solved via the original SVM with two independent minimizations with opposite expected solutions $w^1=-w^0$ and $w_0^1=-w_0^0$. It is worth mentioning that SVM is referred to $\pm 1$ labels thus to compute the regressed observations we need to evaluate:
\begin{equation}\label{eq:1ap}
\hat{\mathbf{Y}}=(\mathbf{X}\mathbf{w}^T(\mathbf{w}\mathbf{w}^T)^{-1}+1)/2
\end{equation}

\bibliographystyle{srt}

\begin{thebibliography}{1}


\bibitem{Cohen2011}
Cohen, J.R.,  et al.  Decoding continuous behavioral variables from neuroimaging
data. Front. Neurosci. 5. 2011.

\bibitem{Reiss15}
Reiss, P.T. et al. Cross-validation and hypothesis testing in neuroimaging: an irenic comment on the exchange between Friston and Lindquist et al. Neuroimage. 2015 August 1; 116: 248–254

\bibitem{Friston2013}
Friston, K. Sample size and the fallacies of classical inference. NeuroImage 81 (2013) 503–504

\bibitem{Bzdok17}
Bzdok, D. Classical Statistics and Statistical Learning in Imaging Neuroscience. Front. Neurosci., 06 October 2017 | https://doi.org/10.3389/fnins.2017.00543

\bibitem{Heller07}
Heller, R. et al. 2007. Conjunction group analysis: An alternative to mixed/random effect analysis. NeuroImage 37, 1178-1185.

\bibitem{Rosenblatt14}
Rosenblatt, J. D.  et al. Revisiting multi-subject random effects in fMRI: Advocating prevalence estimation. NeuroImage 84 (2014): 113-121.

\bibitem{Friston02}
Friston, K.J. et al. Classical and Bayesian inference in neuroimaging: theory NeuroImage, 16 (2) (2002), pp. 465-483

\bibitem{Hastie2001}
Hastie T. et al. The elements of statistical learning theory. Data Mining inference and prediction. Ed Springer. isbn 0-387-95284-5. 2001

\bibitem{Lindquist13}
Lindquist, M.A. et al. Ironing out the statistical wrinkles in "ten ironic rules". Neuroimage. 2013 Nov 1;81:499-502.

\bibitem{Winkler16}
Winkler, A.M., et al. Non‐parametric combination and related permutation tests for neuroimaging. Human brain mapping 37.4 (2016): 1486-1511.

\bibitem{Kohavi95}
Kohavi, R. A study of CV and bootstrap for accuracy estimation
and model selection. Proc. of the 14th international joint
conference on AI - Vol. 2 pp 1137-1143 (1995)

\bibitem{Gorriz18}
Górriz, J.M. et al. A Machine Learning Approach to Reveal the NeuroPhenotypes of Autisms. International journal of neural systems, 1850058. 2019.

\bibitem{Gorriz19}
Górriz, J.M. et al. On the computation of distribution-free performance bounds: Application to small sample sizes in neuroimaging. Pattern Recognition 93, 1-13, 2019.

\bibitem{Varoquaux18}
Varoquaux, G. Cross-validation failure: Small sample sizes lead to large error bars. NeuroImage 180 (2018) 68–77.

\bibitem{Rosenblatt16} 
Rosenblatt, J.D. et al. Better-than-chance classification for signal detection. Biostatistics (2016).

\bibitem{Kim20} 
Kim, I. et al. Classification accuracy as a proxy for two sample testing  Annals of Stat., 2020

\bibitem{Vapnik82}
Vapnik, V. Estimation dependencies based on Empirical Data.
Springer-Verlach. 1982 ISBN 0-387-90733-5

\bibitem{Gorriz2021}
JM Gorriz et al.  Statistical Agnostic Mapping: A framework in neuroimaging based on concentration inequalities.  Information Fusion Volume 66, February 2021, Pages 198-212

\bibitem{Ojala2010}
Ojala M, et al.  Permutation tests for studying classifier performance. Journal of Machine Learning Research. 2010; 11:1833–1863.

\bibitem{Woolrich2009}
Woolrich, et al.  Bayesian analysis of neuroimaging data in FSL.  NeuroImage 45 (2009) S173–S186.

\bibitem{Haussler92}
Haussler, D. Decision theoretic generalizations of the PAC model for neural net and other learning applications. Information and Computation
Volume 100, Issue 1, September 1992, Pages 78-150

\bibitem{Friston12}
Friston, K. Ten ironic rules for non-statistical reviewers.  NeuroImage 61 (2012) 1300–1310

\bibitem{Cover65}
Cover, T.M. Geometrical and Statistical properties of systems of
linear inequalities with applications in pattern recognition. IEEE
Transactions on Electronic Computers. EC-14: 326–334 (1965)

\bibitem{Mcintosh96}
McIntosh, A.R. et al. Spatial pattern analysis of functional brain images using partial least squares, Neuroimage 3(3 Pt 1) (1996) 143-157

\bibitem{Rosipal06}
Rosipal, R. et al. Overview and Recent Advances in Partial Least Squares (Springer Berlin, Heidelberg, 2006), pp. 34-51

\bibitem{Rondina15}
Rondina, J.M. SCoRS - a Method Based on Stability for Feature Selection and Mapping in Neuroimaging. IEEE Trans Med Imaging. 2014 Jan; 33(1): 85–98.

\bibitem{Martinez19}
Martinez-Murcia, F.J. et al. Studying the Manifold Structure of Alzheimer's Disease: A Deep Learning Approach Using Convolutional Autoencoders. IEEE J Biomed Health Inform. 2019 Jun 17.

\bibitem{DeMartimo08}
De Martino, F. et al. Combining multivariate voxel selection and support vector machines for mapping and classification of fMRI spatial patterns
NeuroImage, 43 (1) (2008), pp. 44-58

\bibitem{Vapnik71}
Vapnik, V. et al. On the uniform convergence of relative frequencies of events to their probabilities. Theory of Probability and Its Applications, 16:264–280, 1971.

\bibitem{Massart00}
Massart, P. Some applications of concentration inequalities to
statistics. Annales de la Faculté des Sciences de Toulouse, 2000.

\bibitem{McDiarmid89}
McDiarmid, C. On the method of bounded differences, Surveys in
Combinatorics 141 (1989), 148–188

\bibitem{Sauer72}
Sauer, N. On the density of families of sets. Journal of Combinatorial Theory, Series A, 13:145–147, 1972.

\bibitem{Shela72}
Shelah, S. A combinatorial problem: stability and order for models and theories in infinity languages. Pacific Journal of Mathematics, 41:247–261, 1972.

\bibitem{Gomez19}
Gómez-Verdejo, V. et al. Sign-Consistency Based Variable Importance for Machine Learning in Brain Imaging Neuroinformatics October 2019, Volume 17, Issue 4, pp 593–609

\bibitem{Khundrakpam15}
Khundrakpam, B.S. et al. (2015). Prediction of brain maturity based on cortical thickness at different spatial resolutions. NeuroImage, 111, 350–359.

\bibitem{Mouro-Miranda05}
Mouro-Miranda, J. et al. Classifying brain states and determining the discriminating activation patterns: Support vector machine on functional MRI data. NeuroImage, 28, 980–995. (2005).

\bibitem{Friston95}
Friston, K. et al. Statistical Parametric Maps in functional imaging: A general linear approach Hum. Brain Mapp. 2:189-210 (1995)

%\bibitem{Friston03}
%Friston, K.J., et al. Dynamic causal modelling. Neuroimage 19, 1273–1302, 2003.

\bibitem{Tzourio02}
Tzourio-Mazoyer, N. et al.. Automated anatomical labeling of activations in spm using a macroscopic anatomical parcellation of the MNI MRI single subject brain. Neuroimage 2002; 15: 273-289. DOI

\bibitem{Shalev-Shwartz14}
Shalev-Shwartz, S. et al. Understanding Machine Learning – from Theory to Algorithms. Cambridge University Press. ISBN 9781107057135. 2014

\bibitem{Antos02}
Antós, A. et al. Data-dependent margin-based generalization bounds for classification. Journal of Machine Learning Research 3 (2002) 73–98

\bibitem{Vidyasagar03}
Vidyasagar, M. Learning and Generalisation With Applications to Neural Networks- Springer. ISBN 978-1-84996-867-6 (2003)

\bibitem{Frackowiak04}
Frackowiak, R.S. J. et al. Human Brain Function (Second Edition). Chap. 44. Introduction to Random Field Theory. ISBN 978-0-12-264841-0 Academic Press. 867-879, 2004.

\bibitem{NIA18}
Jack, Jr. C.C. et al. NIA-AA Research Framework: Toward a biological definition of Alzheimer’s disease. Alzheimers Dement. 2018 Apr; 14(4): 535–562.

\bibitem{McKhann11}
McKhann, G.M. et al. The diagnosis of dementia due to Alzheimer’s disease: recommendations from the National Institute on Aging and the Alzheimer’s Assocation Workgroup. Alzheimers Dement. 2011;7:263–9.

\bibitem{Button13}
Button, K.S. et al. Confidence and precision increase with high statistical power Nature Reviews Neuroscience volume 14, page 585(2013).

\bibitem{Illan12}
Illan, I.A. et al. Automatic assistance to Parkinson's disease diagnosis in DaTSCAN SPECT imaging. Medical Physics. 2012

\bibitem{Zaidi06}
Zaidi, H. et al. Quantitative Analysis in Nuclear Medicine Imaging
Springer Science Business Media, Inc. ISBN-10: 0-387-23854-9

\bibitem{Ioannidis05}
Ioannidis, J.P.A. Why most published research findings are false. PLoS Med. 2 (8) (e124), 696–701. 2005.

\bibitem{Eklund16}
Eklund, A. et al. Cluster failure: Inflated false positives for fMRI. Proceedings of the National Academy of Sciences Jul 2016, 113 (28) 7900-7905; DOI: 10.1073/pnas.1602413113

\bibitem{Lorca18}
Lorca-Puls, D.L. et al. The impact of sample size on the reproducibility of voxel-based lesion-deficit mappings, Neuropsychologia, Volume 115,
2018,101-111, 2018

\bibitem{Abraham14}
Abraham, A. et al. Machine learning for neuroimaging with scikit-learn. Frontiers in Neuroinformatics. Vol 8. pages 14. 2014.

\bibitem{Gareth98}
Gareth, J. et al. The error coding method and picts. Journal of Computational and Graphical Statistics, 7(3):377–387, 1998.

\bibitem{Gorriz2020}
Gorriz, J.M. et al. Artificial intelligence within the interplay between natural and artificial computation: Advances in data science, trends and applications. Neurocomputing Volume 410, 14 October 2020, Pages 237-270

\bibitem{Allefeld16}
Allefeld, C. et al. Valid population inference for information-based imaging: From the second-level t-test to prevalence inference. Neuroimage 141 (2016): 378-392.

\bibitem{Alain08}
Rakotomamonjy, A. et al. SimpleMKL Journal of Machine Learning Research 9 (Nov), 2491-2521. 2008

\bibitem{Ashburner05}
Ashburner, J. et al. Unified segmentation, NeuroImage,Volume 26, Issue 3, Pages 839-851, 2005

\bibitem{Burges98}
Burges, C.J.C. A tutorial on support vector machines for pattern recognition
Data Mining and Knowledge Discovery, 2 (2) (1998), pp. 121-167

\end{thebibliography}

\end{document}